\documentclass[acmtog]{acmart} 
\usepackage{algorithm}
\usepackage{bm}
\usepackage{algpseudocode}
\usepackage{graphicx}
\usepackage{subcaption}
\usepackage{tabularx}
\usepackage{diagbox}
\usepackage{multirow}
\usepackage{booktabs}
\usepackage{amsmath,amsopn,amstext,amsfonts}
\usepackage{pifont}
\usepackage{bbding}
\usepackage{stfloats}
\usepackage{xcolor}
\usepackage{tikz}
\usepackage[utf8]{inputenc}
\DeclareUnicodeCharacter{FE20}{\textlangle} 
\DeclareUnicodeCharacter{FE21}{\textrangle} 

\AtBeginDocument{%
  \providecommand\BibTeX{{%
    \normalfont B\kern-0.5em{\scshape i\kern-0.25em b}\kern-0.8em\TeX}}}

\setcopyright{acmlicensed}
\acmJournal{TOG}
\acmYear{2024} \acmVolume{43} \acmNumber{4} \acmArticle{110} \acmMonth{7}\acmDOI{10.1145/3658181}

\begin{document}

\title{Implicit Swept Volume SDF: Enabling Continuous Collision-Free Trajectory Generation for Arbitrary Shapes}

\author{Jingping Wang}
\authornote{Both authors contributed equally to this research. \\
$^{\dag \,}$Corresponding authors.}
\email{22232111@zju.edu.cn}
\orcid{0009-0001-2865-6517}
\author{Tingrui Zhang}
\authornotemark[1]
\email{tingruizhang@zju.edu.cn}
\orcid{0000-0003-2130-8226}
\affiliation{%
  \institution{Zhejiang University}
  \city{Hangzhou}
  \country{China}
  \postcode{310013}
}

\author{Qixuan Zhang}
\affiliation{%
  \institution{ShanghaiTech University and Deemos Technology Co., Ltd.}
  \city{Shanghai}
  \country{China}}

\author{Chuxiao Zeng}
\affiliation{%
  \institution{ShanghaiTech University and Deemos Technology Co., Ltd.}
  \city{Shanghai}
  \country{China}}

\author{Jingyi Yu}
\affiliation{%
  \institution{ShanghaiTech University}
  \city{Hangzhou}
  \country{China}}
  
\author{Chao Xu}
\affiliation{%
  \institution{Zhejiang University}
  \city{Hangzhou}
  \country{China}}

\author{Lan Xu$^{\dag \,}$}
\affiliation{%
  \institution{ShanghaiTech University}
  \city{Shanghai}
  \country{China}}
  
\author{Fei Gao$^{\dag \,}$}
\email{fgaoaa@zju.edu.cn}
\affiliation{%
  \institution{Zhejiang University}
  \city{Hangzhou}
  \country{China}}



\renewcommand{\shortauthors}{Wang and Zhang, et al.}
\begin{teaserfigure}
  \includegraphics[width=1.0\textwidth]{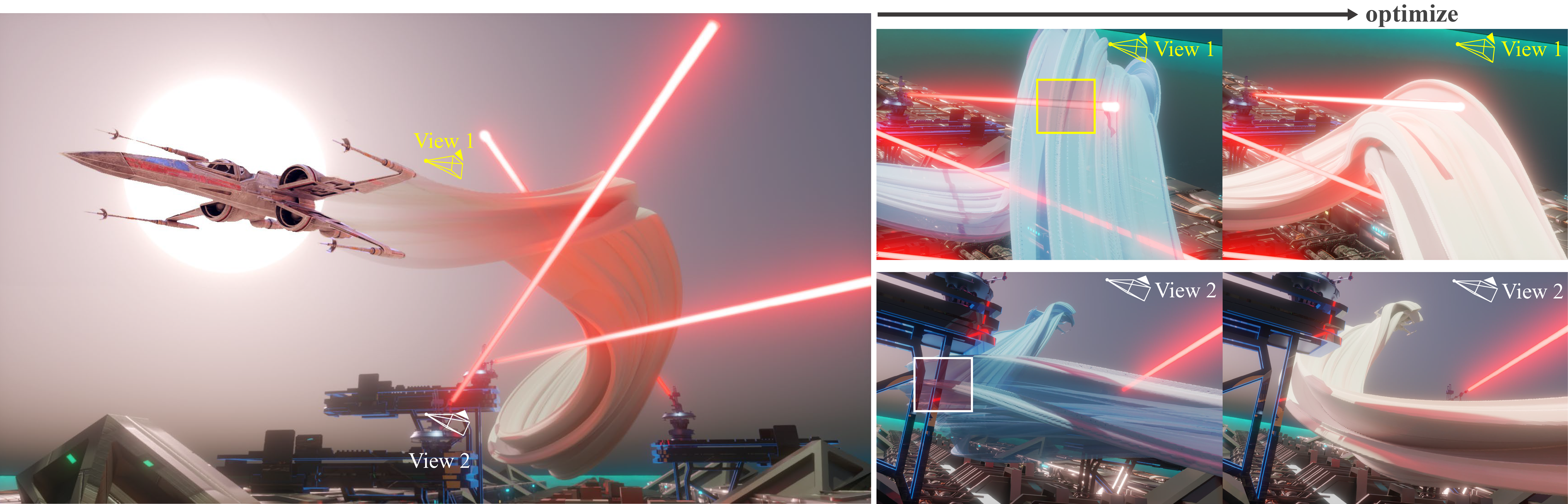}
  \captionsetup{font={small}}
  \captionof{figure}{ \label{fig: top} Our approach facilitates the generation of continuous, collision-free animated trajectories for the X-wing. The left figure shows the swept volume of the X-wing during flight. The four figures on the right show the trajectories before and after optimization from two different perspectives. The motions are visualized using swept volumes in the figure, providing an intuitive representation of continuous collision scenarios.
  }
  \label{fig:head}
\end{teaserfigure}

\begin{abstract}
In the field of trajectory generation for objects, ensuring continuous collision-free motion remains a huge challenge, especially for non-convex geometries and complex environments.
Previous methods either oversimplify object shapes, which results in a sacrifice of feasible space or rely on discrete sampling, which suffers from the ``tunnel effect''.
To address these limitations, we propose a novel hierarchical trajectory generation pipeline, which utilizes the \textbf{S}wept \textbf{V}olume \textbf{S}igned \textbf{D}istance \textbf{F}ield (SVSDF) to guide trajectory optimization for \textbf{C}ontinuous \textbf{C}ollision \textbf{A}voidance (CCA).
Our interdisciplinary approach, blending techniques from graphics and robotics, exhibits outstanding effectiveness in solving this problem.
We formulate the computation of the SVSDF as a Generalized Semi-Infinite Programming model, and we solve for the numerical solutions at query points implicitly, thereby eliminating the need for explicit reconstruction of the surface.
Our algorithm has been validated in a variety of complex scenarios and applies to robots of various dynamics, including both rigid and deformable shapes. It demonstrates exceptional universality and superior CCA performance compared to typical algorithms. 
The code will be released at \textcolor{red!70}{\url{https://github.com/ZJU-FAST-Lab/Implicit-SVSDF-Planner}}  for the benefit of the community.

\end{abstract}




\maketitle
\section{Introduction}
Generating continuous collision-free motion trajectories for objects of any shape is highly valuable in fields like animation production, computer-aided design, manufacturing, and robotic navigation planning.
However, in practical applications, the shapes of objects and their environments are often complex, non-convex geometries.
This makes handling collisions during continuous motion a challenge.
Previous methods have either oversimplified the shapes of objects and environments or approximated continuous motion with discrete sampling moments.
The former compromises feasible space, leading to an inability to generate correct motion trajectories in complex, confined environments.
The latter theoretically risks missing collision detections, failing to ensure continuous collision-free motion. This phenomenon is known as the ``tunnel effect'' \cite{ericson2004real}.
Due to these challenges, achieving \textbf{C}ontinuous \textbf{C}ollision \textbf{A}voidance (CCA) without simplifying the shapes of objects or sacrificing any feasible space has been a long-pursued goal in trajectory generation algorithms.

Inspired by recent advances in computer graphics and robotics, we propose a new perspective. 
The \textbf{S}wept \textbf{V}olume (SV) generated by an object's continuous motion describes the minimal safe space required throughout its movement.
This means that if there are no obstacles inside the SV created by an object's motion, then continuous collision-free motion is theoretically guaranteed.
Moreover, this assurance does not require simplifying the shapes of objects or environments.
From this viewpoint, we introduce a novel pipeline for generating continuous, collision-free trajectories for objects of arbitrary shape. Our approach relies on the numerical optimization of spatiotemporal joint trajectories.
Using the implicit \textbf{S}igned \textbf{D}istance \textbf{F}ield (SDF) of the SV, we can quickly generate gradients at obstacles that pose  collision risks,
which guides trajectory optimization.
Owing to the inherent compactness of the SV in describing the occupied space, our method naturally achieves zero feasible space sacrifice and CCA.
Moreover, our method does not suffer from the  ``tunnel effect'' \cite{ericson2004real}.
To the best of our knowledge, this is the first method to concurrently achieve these outcomes in motion trajectory generation.

We first propose a method to compute the implicit SDF of the SV.
In previous research, much focus has been on the reconstruction of the surface of the SV.
However, the task of surface reconstruction itself is extremely challenging.
Simply obtaining the surface area of the SV does not provide a differentiable objective function for trajectory optimization either.
In trajectory generation tasks, we are more concerned with the signed distance of obstacles relative to the SV.
A smaller signed distance value means that the obstacle is closer to the object at a certain moment.
If the sign is negative, it indicates an impending collision with the obstacle, which is a scenario to be avoided.
In this paper, we formulate the problem of solving for the \textbf{S}wept \textbf{V}olume \textbf{S}igned \textbf{D}istance \textbf{F}ield (SVSDF) as a \textbf{G}eneralized \textbf{S}emi-\textbf{I}nfinite \textbf{P}rogramming (GSIP) model.
We propose a method to implicitly solve the swept volume signed distances at query points without explicitly reconstructing the SV surface.
Theoretically, our method can compute the exact SVSDF and can be proven to converge to any numerical precision.
We then developed a hierarchical trajectory optimization algorithm.
This algorithm uses the implicit SVSDF to guide the SV away from obstacles.
Additionally, our approach optimizes the energy of the trajectory, such as minimizing the integral of the squared control efforts in robotic planning tasks.

Our algorithm, proven effective across diverse scenarios with complex-shaped vehicles, aircraft, ships, and deformable worm and ferrofluid robots, demonstrates its universality. It also shows superior CCA performance compared to previous methods.

The contributions can be summarized as follows:
\begin{itemize}
\item We propose a novel method based on GSIP to compute the exact SVSDF of various shapes.
\item We develop a trajectory generation framework centered on hierarchical optimization that enables continuous collision safety for arbitrarily shaped robots.
\item Our algorithm demonstrates state-of-the-art CCA performance in a variety of scenarios, showcasing its versatility and broad adaptability.
\item We will open-source our algorithms to support the graphics 
 and the robotics community.
\end{itemize}

\section{Related Works} 
\subsection{Swept Volume SDF Calculation}

There is a long history of research on SV.
Over the past decades, numerous works have contributed to the techniques for computing SV. However, previous efforts have focused more on constructing the surfaces of SV, such as methods based on envelope theory \cite{wang1986geometric,martin1990sweeping,weld1990geometric}, methods utilizing differential equations\cite{blackmore1997sweep}, kinematic methods \cite{ju1996computer}and methods based on exact Boolean calculations \cite{zhou2016mesh,cherchi2020fast}. 

Recently, a novel approach was proposed by \citet{sellan2021swept} that extends the zero-level set of spatiotemporally continuous implicit functions to obtain the surface of SV. This approach has significant improvements in generalization, robustness, and efficiency. The proposed implicit function is derived from the minimum signed distance during the brush motion process. As part of the method, the implicit function itself is a conservative SDF (cf. \cite{iquilezles2018,sellan2023reach,sellan2021swept}) of the SV, and has exact values only outside the SV. 
However, in the trajectory generation process, the SDF inside SV is more critical, as it guides the SV to avoid obstacles that could lead to collisions.

The first work to achieve exact SDF computation for SV is presented in \cite{marschner2023constructive}. The authors use the closest point loss to correct for conservative SDFs generated by previous methods using neural networks in the context of the CSG operation. By adding a loss function to fit the SV, the network can obtain the exact SDF of the SV generated under a cubic Bézier path. However, neural networks require hours of training for accurate SDF evaluation. Furthermore, encoding different trajectories and shapes is challenging, and it often requires additional training or adjustments to the input dimensions of the network encoder. In addition, the output of neural networks lacks precise theoretical guarantees. In contrast, our numerical method computes the exact SVSDF under theoretical guarantees and applies to different trajectory shapes and brushes, requiring only the SDF of the brushes.

\subsection{Continuous Collision Avoidance Trajectory Generation}
In robotics and graphics, the generation of continuous, collision-free motion has become a focal point. Early research focused on path generation, which identifies point sequences 
while ensuring segment safety, using either search-based (e.g. \cite{hart1968formal,frana2010interview}) or sampling-based (e.g. \cite{janson2015fast,lavalle1998rapidly}) methods. 
However, these methods struggle with resolution sensitivity, challenges in unstructured environments, and satisfying complex nonlinear constraints such as dynamic constraints. 
Recently, the focus has shifted to accurately representing continuous motion trajectories and optimizing them through numerical optimization, 
a strategy that is adept at dealing with nonlinear constraints and producing higher quality solutions.

Most of the methods perform collision evaluations at discrete states along the trajectory during optimization. Discrete approximations of continuous-time problems theoretically hinder CCA. This limitation arises because the discretization of the states potentially leads to the underdetection of collisions.  For example, in a game scenario, a fast-moving projectile may pass through a thin wall if the collision is not detected at a certain timestamp.
This phenomenon is known as the  ``tunnel effect'' \cite{ericson2004real} and is a challenge that Continuous Collision Detection algorithms aim to address. Continuous Collision Detection represents a group of algorithms designed for robust collision check continuously, rather than just at discrete moments \cite{brochu2012efficient,wang2021large}. However, these technologies are not directly applicable to trajectory generation because they are limited to providing Boolean detection results and cannot provide guidance or gradients to optimize collision-free trajectories.

Many efforts are dedicated to achieving CCA. Safe corridor based methods divide safe areas into convex hulls and restrict trajectories within these hulls \cite{7839930,8740885}. 
However, these methods compromise the feasible space and are not suitable for complex shapes or dense obstacles. The work  \cite{blackmore1992analysis} presents a mathematical technique for analyzing SV and shows that the notion of a sweep differential equation leads to criteria that provide useful insights into the geometric and topological properties of SV, although these insights have not been applied to motion planning for robots. In the work  \cite{guthrie2022differentiable}, the concept of SV is used to solve the CCA problem, but the SV is only approximated by convex hulls, which is not tight, and the application of the robots is only for planar two-dimensional cars, which cannot be applied to robots of arbitrary shapes. \citet{hauser2021semi} focus on collision constraints for non-convex geometries in robot planning. They compute the maximum penetration depth between two rigid bodies and effectively address these constraints using semi-infinite programming. However, their approach is based on a discrete branch-and-bound method that neglects the continuity of object motion. Recently, \citet{10342104} attempt to address CCA for robots using trajectory optimization based on implicit signed distance fields.
However, their approach lacks a comprehensive planning framework and fails to correctly compute the SDF inside the SV during optimization, resulting in gradient oscillations. In complex scenarios, their success rate in achieving CCA decreases.

In conclusion, no existing planning algorithm has been able to achieve effective CCA without compromising the solution space. To address this problem, our method pioneers the use of SVSDF combined with advanced hierarchical optimization techniques, resulting in unprecedented performance. 
Specifically, we solve a GSIP to compute the exact SVSDF. 
With this exact SVSDF, we establish a CCA framework for objects of any shape, ensuring tight and reliable continuous collision estimation applicable to trajectory generation. 
Our method has been benchmarked and achieved the highest level of CCA performance to date.
\section{Implicit Swept Volume SDF}
In this chapter, we first introduce the method for computing SVSDF. By modeling the problem as a GSIP, our method can compute an exact SVSDF over the entire space, which provides accurate guidance for generating trajectories of arbitrary shapes in complex environments.

We denote the mathematical concept of the SV set with the symbol $\mathcal{SV}$, and the italicized $SVSDF(\bm{p})$ represents the SVSDF query function that returns the SVSDF value at the point $\bm{p}$.

\subsection{GSIP Model for Swept Volume SDF}
\label{GSIP Concept}

$\mathcal{SV}$ refers to the set of all points that an object passes through as it moves along a trajectory.
The set $M(t)$ is used in this paper to represent potentially time-varying shapes. Suppose $M(t)$ moves along the trajectory 
$\mathcal{T}(t)$, where $t$ is the time parameter, $\mathcal{SV}$ can be defined as
\begin{align}
\mathcal{SV} = \bigcup_{t \in [t_{\text{start}}, t_{\text{end}}]}   \ \mathcal{T}(t)M(t).
\end{align}

\begin{figure}[!tb] 
\centering
{\includegraphics[width=0.85\columnwidth]{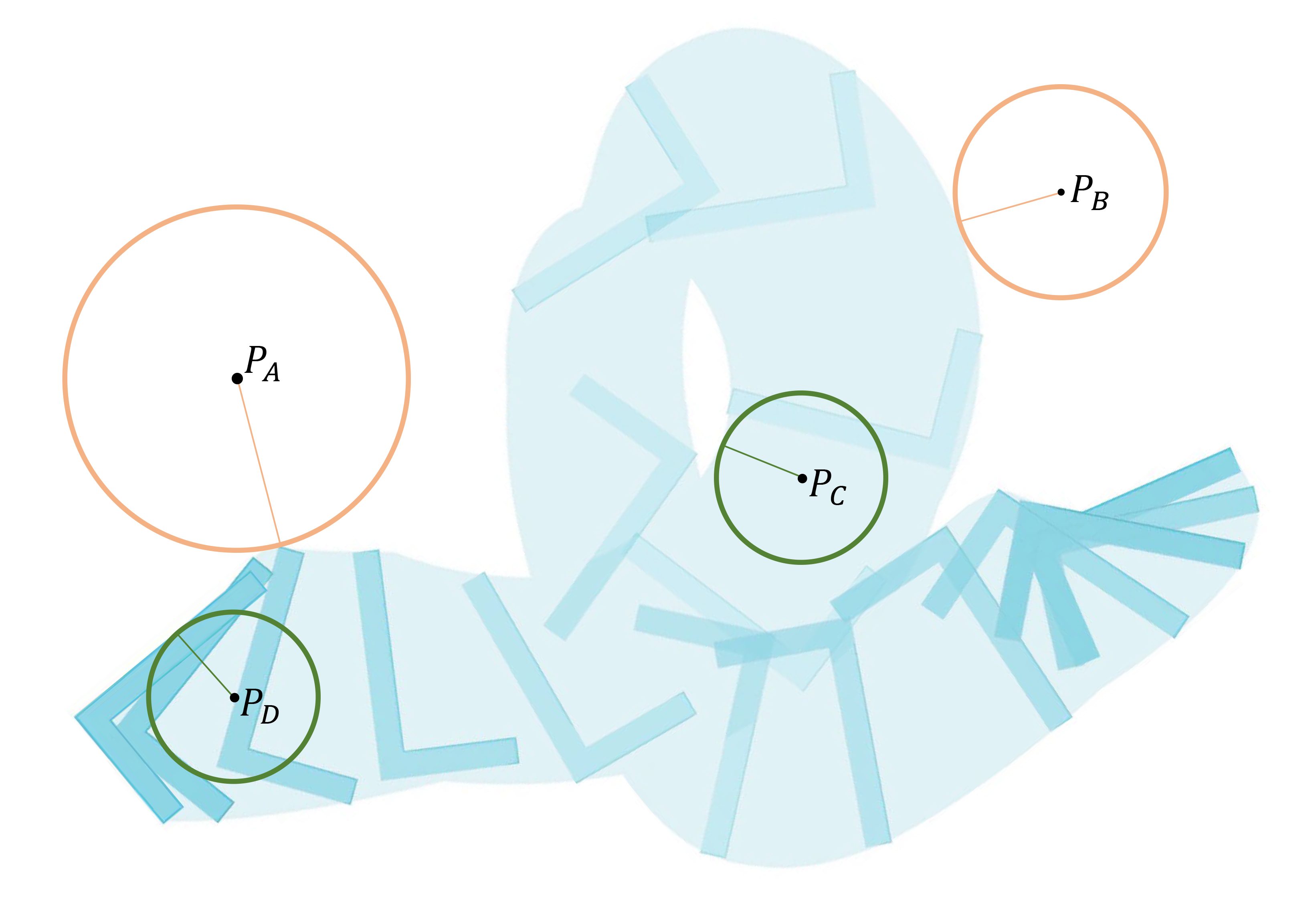}}
\captionsetup{font={small}}
\caption{
\label{fig: svsdf_a}  
An L-shaped object follows a path that combines translation and rotation, creating the light blue SV. Points $\bm{p}_A$ and $\bm{p}_B$ are outside the SV, marked by yellow circles indicating the smallest tangent circles with radii equal to the SVSDF values at those points. In contrast, points $\bm{p}_C$ and $\bm{p}_D$ are inside the SV, with their SVSDF values corresponding to the negative radii value of the centered green circles.}
\end{figure}

In the formula mentioned above, we use the homogeneous rigid transformation matrix $\mathcal{T}$ to represent the trajectory. Thus, $\mathcal{T}(t)M(t)$ intuitively means the position and pose of shape $M(t)$ at time $t$. Computing the SVSDF is essentially a matter of finding the shortest signed distance metric from a point $\bm{p}$ to $Fr(\mathcal{SV})$, where the symbol $Fr(\cdot)$ denotes the boundary of a set.
Consider an open ball $B_{\bm{p}}(r)$ centered at point $\bm{p}$ with radius $r$. Intuitively, if $B_{\bm{p}}(r)$ is the smallest sphere at $\bm{p}$ that tangent to $Fr(\mathcal{SV})$, then $|SVSDF(\bm{p})| = r$. Fig. \ref{fig: svsdf_a} illustrates this property.


Inspired by this feature, our algorithm calculates the SVSDF magnitude at point $\bm{p}$ by solving for the sphere centered at $\bm{p}$ that is tangent to $Fr(\mathcal{SV})$. Specifically, we solve the following problem:

\begin{equation}
\begin{aligned}
\label{equ:maxr}
    \text{maximize} \quad   & r, \\
    \text{s.t.} \quad B_{\bm{p}}(r) \cap Fr(\mathcal{SV}) & = \emptyset.  
\end{aligned}
\end{equation}

However, the constraint that two infinite sets do not intersect is very difficult to deal with, but fortunately, this constraint can be reformulated into infinite inequality constraints, thus transforming the problem into a standard GSIP in mathematical optimization.

Specifically, if there is a metric function $g$, which maps a point $\bm{q}$ to a real value, $g(\bm{q})$ > 0 when $\bm{q}$ lies outside the swept volume and $g(\bm{q})$ < 0 when $\bm{q}$ lies inside the swept volume, then problem (\ref{equ:maxr}) can be equivalently transformed as:

\begin{align}
    & \quad \text{maximize} \quad r, \notag  \\
    \text{s.t.} \quad g(\bm{q}) \leq 0, \ \forall \bm{q} &\in B_{\bm{p}} \quad \text{if} \quad \bm{p} \ \in \ \mathcal{SV}, \label{equ:inside}\\
    \text{s.t.} \quad g(\bm{q}) \geq 0, \ \forall \bm{q} &\in B_{\bm{p}} \quad \text{if} \quad \bm{p} \ \notin \ \mathcal{SV}. \label{equ:outside}
\end{align}

That is, the open ball $B_{\bm{p}}(r)$ is tangent to the $Fr(\mathcal{SV})$ while all the points inside the ball are either outside the $Fr(\mathcal{SV})$ or all inside, depending on whether the centre of the ball is inside the $Fr(\mathcal{SV})$ or not.
In this work, we chose $g$ that satisfies the condition as:

\begin{align}
\label{min_distance}
g(\bm{p}) \! \triangleq \! \underset{t \in [t_{\text{start}}, t_{\text{end}}]}{\operatorname{min}} \! \mathcal{SDF}^{M(t)} \! \left( \mathcal{T}^{-1}(t)\ \bm{p} \right)\!,
\end{align}

\begin{figure*}[!h]  
\vspace{0.0cm}  
\centering
{\includegraphics[width=1.8\columnwidth]{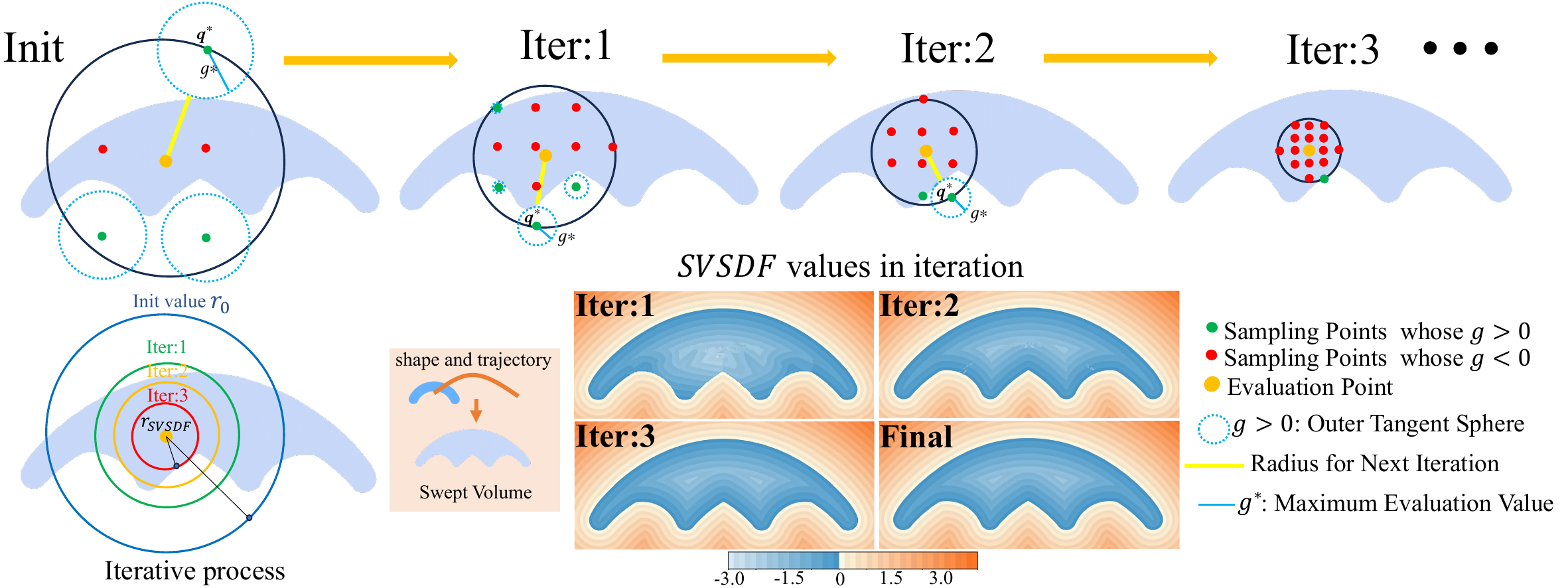}}
\captionsetup{font={small}}
\caption{ \label{fig:real1_1}The figure shows a simplified iterative method for computing the internal SVSDF in 2D using the GSIP, a process that is also applicable in 3D. In the discretized data, green points outside SV use gradient descent (as described in Section \ref{acceleration strategy}) to find their metric value $g$, which is the radius of the tangent circle. The largest $g^{*}$ among these values indicates the largest current constraint violation. The radius of the next iteration is reduced by $g^{*}$. Increasing the sampling density and iterations quickly yields accurate SDF values within SV.}
\vspace{0.00cm}  
\label{fig: iteration_vis}
\end{figure*}

where the associated argmin time is denoted as follows:
 \begin{align}
 \label{argmin_t}
\setlength\abovedisplayskip{5pt}
\setlength\belowdisplayskip{5pt}
     &t^{*}(\bm{p}) \! \triangleq \! \underset{t \in\left[t_{\text{start}}, t_{\text{end}}\right]}{\operatorname{argmin}} \! \mathcal{SDF}^{M(t)} \! \left(\mathcal{T}^{-1}(t)\ \bm{p}\right)\!.
\end{align}

Here, $\mathcal{SDF}^{M(t)}$ represents the SDF of the shape $M(t)$, and $\mathcal{T}^{-1}(t)\ \bm{p}$ is the relative position of the point $\bm{p}$ in the coordinate system of $M$.
In fact, the work \cite{sellan2021swept} has proved that the function $g$ defined in this way is a conservative SDF of the $\mathcal{SV}$. i.e., if $\bm{q}$ lies outside the $\mathcal{SV}$, then $g(\bm{q}) = SVSDF(\bm{q})$. If $\bm{q}$ lies inside, then $g(\bm{q})$ is the upper bound of $SVSDF(\bm{q})$.
This choice offers two advantages. First, for the case of $g(\bm{q}) \geq 0$ in Eq. (\ref{equ:outside}), the signed distance can be obtained directly, thus focusing our attention solely on solving Eq. (\ref{equ:inside}). Second, the nature of the conservative SDF contributes to the fast convergence of the optimization variables when solving the GSIP, as will be shown in the Section \ref{Construction of the GSIP}.

\subsection{Solving the GSIP problem}
\label{Construction of the GSIP}

In a general form, GSIP can be stated as:
\begin{equation}
\begin{aligned}
    \label{equ:gsip}
    &\text{minimize} \quad    f(x), \\
    &\text{s.t.} \ x\ \in \  Q, \\
    Q\ =\ \{x\  \in \mathbb{R}^n \ &| \   \tilde{g} (x,y) \leq 0 ,\ \forall y \in Y(x) \}.
\end{aligned}    
\end{equation}
The set-valued mapping $Y: \mathbb{R}^n \Rightarrow \mathbb{R}^m $ describes the index set of inequality constraints. 
Functions $f$ and $\tilde{g}$ in the equation above are assumed to be real-valued and at least continuous on their respective domains \cite{stein2012solve}.
By exploiting the bi-level structure of the GSIP, we can effectively tackle the problem. 

Specifically, we replace the infinite set $Q$ by the finite set $Q^{'}$ in Eq. (\ref{equ:gsip}) by first finding the upper bound of the constraint:

$Q^{'}\ = \{x\ \in \mathbb{R}^n \ | \operatorname{sup}_{{y \in Y(x)}} \tilde{g} (x,y) \leq 0\}$.

This insight decomposes the GSIP into a bi-level optimization problem: 
first solve the low-level problem $\bm{LP(x,y)}$ to find an upper bound for the infinite constraints, and then solve the upper-level problem $\bm{UP(x,y)}$, which solves the optimization problem with finite constraints only.
\begin{align}
\label{equ:lp}
  \bm{LP(x,y):}& \quad y^{*} = \underset{y}{argmax}\ \tilde{g} (x,y)\   ,\  s.t.\ y\ \in \  Y(x), \\
 \bm{UP(x,y):}& \quad \underset{x}{minimize}\ f(x)\   ,\  s.t.\ \tilde{g} (x,y^{*}) \leq 0.
 \label{equ:up}
\end{align}

By substituting Eq. (\ref{equ:inside}) into forms above  we can derive our bi-level formulation: 
\begin{align}
\label{equ: GSIPlp}
  \bm{LP(r,s):}& \quad \bm{s}^{*} = \underset{\bm{s}}{argmax}\ g(\bm{q}(\bm{s}))\   ,\  s.t.\ \bm{q}(\bm{s})\ \in \ Y(r) \equiv B_{\bm{p}}(r) , \\
 \bm{UP(r,s):}& \quad \underset{r}{minimize}\ (-r) \   ,\  s.t.\ g(\bm{q}(\bm{s}^{*})) \leq 0.
 \label{equ:GSIPup}
\end{align}
Here $\bm{s} = \{\theta, \phi, \alpha\}$, where $\theta$ and $\phi$ represent the angles in the spherical coordinate system, and $\alpha$ serves as a scaling factor for the radius $r$, constrained within the range of 0 to 1. The point $\bm{q}(\bm{s})$ is defined in this spherical coordinate system as follows:
\begin{equation}
\label{equ: spherical}
\begin{aligned}
\bm{q}(\bm{s}) &= \bm{p} + [x_r, y_r, z_r]^T, \\
x_r &= \alpha r\sin(\theta)\cos(\phi),\\
y_r &= \alpha r\sin(\theta)\sin(\phi),\\
z_r &= \alpha r\cos(\phi).
\end{aligned}
\end{equation}
where $\bm{p}$ represents the center of the sphere.
However, when $\bm{LP(x,y)}$ and $\bm{UP(x,y)}$ are non-convex problems, it is hard to obtain the optimal solution in one step. 
A more robust approach is to solve iteratively by a discretization method \cite{remez1962general,blankenship1976infinitely}.
In the problem (\ref{equ: GSIPlp}), $\bm{LP(r,s)}$ cannot be guaranteed to be convex, depending on the spatial distribution of the trajectory. Therefore, we adopt the discretization method for numerical solution. Fortunately, the $\bm{UP(r,s)}$ is a linear problem with an analytical solution, simplifying our algorithm. The algorithm procedure is demonstrated in Alg. \ref{adaptive discretization method}, and Fig. \ref{fig: iteration_vis} illustrates the iterative process of the algorithm.
By solving the GSIP, we implicitly obtain the SVSDF value at the query point, without the need for explicit surface reconstruction. The convergence proof of our discretization method is given in detail in $\S$C  of the supplementary materials.

\begin{algorithm}[!h]
\caption{$SVSDF$ Computation}
\label{adaptive discretization method}
\begin{algorithmic}[1] 

\Function{SampleInBall}{$r, \bm{p}$} 
\Comment{}

\State Sample a number of points uniformly inside the ball $B_{\bm{p}}(r)$ 
\State to form a set $Y$ by discretizing $\bm{s}$ in Eq. (\ref{equ: spherical}).
    \State \Return $Y$
\EndFunction
\State --------------------------------------
\State $\textbf{Input:}\ query\ point \ \bm{p}$
\If{ $g(\bm{p}) > 0$ }
    \State \Return $g(\bm{p})$
\Else
    \State $k \leftarrow 0$
    \State $r_k \leftarrow\  a\  big\  initial\  value$. 
    \State $Y_k^{'} \leftarrow \textproc{SampleInBall}(r_k, \bm{p})$
    
    \State $\bm{s}_{k}^{*} \leftarrow$ Replace $Y(r)$ by $Y_k^{'}$ and solve $\bm{LP(r_k,s)}$ in Eq. (\ref{equ: GSIPlp}).
\If{ $g(\bm{q}(\bm{s}_{k}^{*})) < \epsilon_{Numerical Precision}^{+}$ }
    \State \Return $-r_k$
\EndIf
\State Solve $\bm{UP(}\bm{r}_k,\bm{s}_{k}^{*}\bm{)}$ and update $r$, which has an analytic 
\State solution: $r_{k+1} \leftarrow r_k - g(\bm{q}(\bm{s}_{k}^{*}))$.
\State $k \leftarrow k+1$ , $\bm{goto}$ line 13.
\EndIf
\end{algorithmic}
\end{algorithm}

\section{Trajectory Generation with Implicit SVSDF}

In the previous chapter, we introduced the method to compute the implicit SVSDF by solving the GSIP. Thanks to the compactness of the SV, we propose a pipeline based on SVSDF, achieving the trajectory generation method that simultaneously incorporates zero solution space sacrifice and CCA. Our method applies to various configuration spaces, including $\mathbb{R}(2)$, $\mathbb{SE}(2)$, $\mathbb{R}(3)$, and $\mathbb{SE}(3)$, etc., thus satisfying different types of problems. For example, the workspace of ground robots is typically $\mathbb{SE}(2)$, that of drones is $\mathbb{SE}(3)$.
To explain the details of our method, we will use the $\mathbb{SE}(3)$ space and multirotor dynamics \cite{mellinger2011minimum} as a case study in this chapter.
\begin{figure*}[!thb] 
\centering
{\includegraphics[width=1.8\columnwidth]{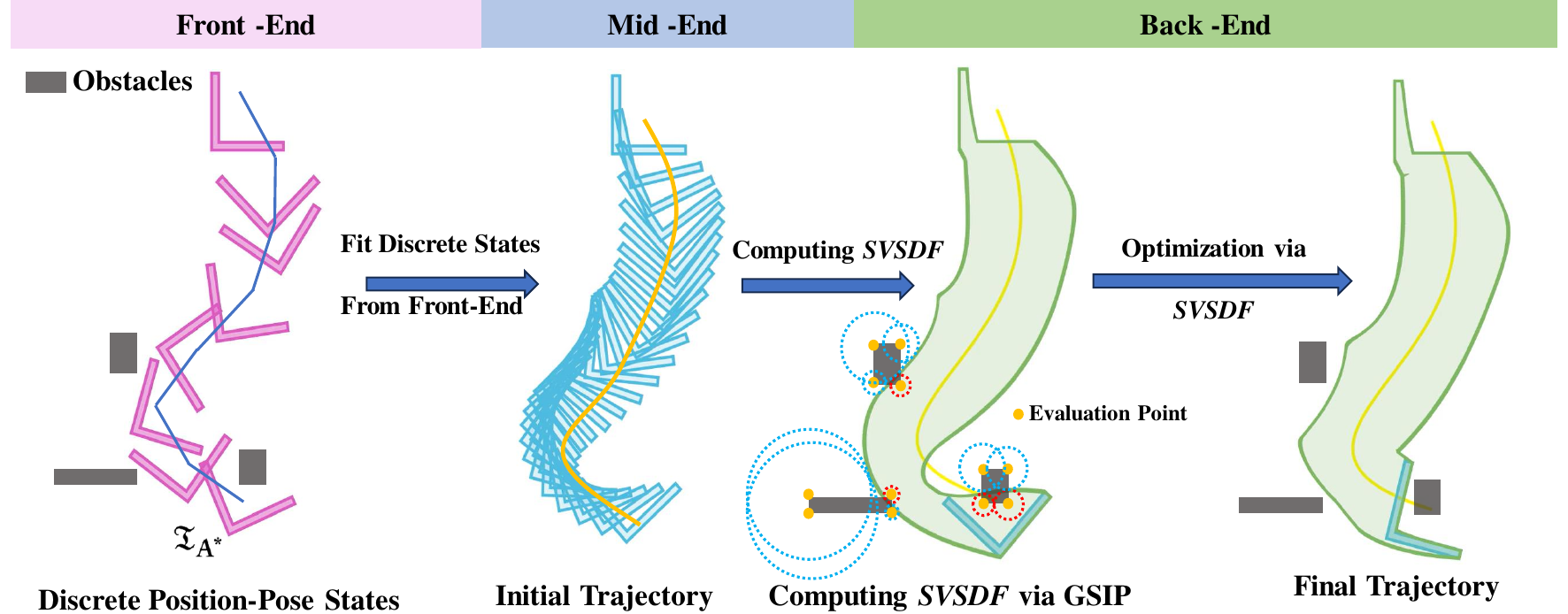}}
\captionsetup{font={small}}
\caption{ \label{fig: hierarchical schematic} The hierarchical trajectory generation framework consists of three stages: the front-end, the mid-end, and the back-end. The front-end generates a discrete sequence of high-dimensional position-pose states, the mid-end facilitates the generation of initial values for the optimized trajectory, and the back-end uses the exact SVSDF to formulate a continuous collision-avoidance trajectory.}
\end{figure*} 
\subsection{Hierarchical Trajectory Optimization}
\label{Hierarchical Trajectory Optimization}

In $\mathbb{SE}(3)$, the configuration of the rigid object at time $t$ is determined by both the rotation matrix $R(t)$ and the translation vector $p(t)$, as expressed by the equation: $\mathcal{T}(t)M(t) = R(t)M(t) + p(t)$.
Compared to sampling and search methods, numerical optimization offers the advantage of incorporating dynamics, making it a prevalent and effective approach for handling high-dimensional motion planning \cite{latombe2012robot}.

However, achieving collision-free trajectory generation in complex environments has an obvious non-convex nature \cite{liu2018convex}, especially when dealing with non-convex shapes without simplification. In such cases, the non-convex nature of the CCA problem makes it challenging for optimization methods to find feasible optimal solutions.
Therefore, numerical optimization methods often adopt a hierarchical planning approach, necessitating the use of sampling-based or search-based methods to provide an initial value for numerical optimization. 
In this paper, the trajectory generation based on SVSDF that we propose is also a hierarchical method based on numerical optimization. Specifically, our proposed pipeline includes three levels:
\begin{itemize}
    \item [1)] \textbf{Front-End}, considering objects of any shape, we use rapid collision detection technology and asymmetric A* search to quickly find a feasible path.
    \item [2)] \textbf{Mid-End}, the feasible path points, including position and attitude, are converted into parameters of a continuous trajectory, obtaining initial values for the trajectory parameters.
    \item [3)]  \textbf{Back-End}, utilizing SVSDF and the initial values provided by the mid-end for trajectory optimization, to achieve CCA and dynamic constraints. 
\end{itemize}
Fig. \ref{fig: hierarchical schematic} illustrates our hierarchical pipeline. \\
\textbf{Front-End}
\label{Front End} \\
Given the start and end points of the planning, obtaining the parameters for a continuous collision-free trajectory through optimization is a highly non-convex problem. Therefore, a good initial value for optimization is required to ensure that the optimization results reach a satisfactory local optimum. In the front-end step, we find a rough feasible path through A* path search in the workspace, guiding the final topological structure of the trajectory.
However, the direct application of vanilla A* faces some challenges, as planning in C-space \cite{619371} requires extensive collision checks and node search expansions. This can be extremely slow in high-dimensional spaces like $\mathbb{SE}(3)$ \cite{ding2019efficient}. 
Therefore, we have adapted the A* method to facilitate efficient path searching within the  $\mathbb{SE}(3)$ space. This adaptation ensures that the search space complexity of our modified A* algorithm aligns with that of the conventional 3D vanilla A*.

\begin{figure}[!t] 
\centering
{\includegraphics[width=0.9\columnwidth]{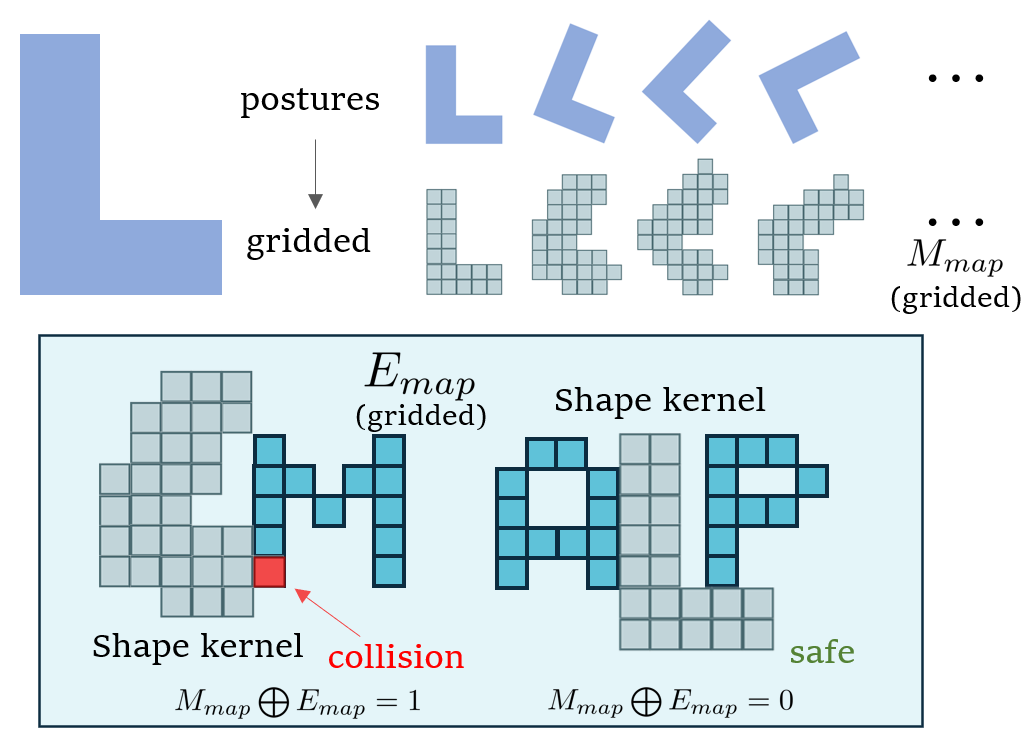}}
\captionsetup{font={small}}
\caption{ \label{fig: front-end}  Here is a 2D example for visual clarity: An L-shaped robot performs collision detection with the environment, where the pose dimension is discretized by the yaw.}
\end{figure}

Our modifications mainly involve two aspects. Firstly, when expanding neighboring nodes, we only evaluate nodes adjacent to the current node in the position dimensions, while in the attitude dimensions, we directly find the feasible attitudes closest to the attitude of the current node, making the expansion of neighboring nodes asymmetric in different dimensions. In trajectory optimization, the continuity of attitudes will be ensured through the energy optimality of the trajectory. Secondly, we utilize discretized collision detection instead of accurate but costly geometric collision detection, which can be visualized with the help of Fig. \ref{fig: front-end}. In fact, the final trajectory is optimized according to the SVSDF, thus the front-end path search does not need to be overly fine and strictly collision-free. The use of precise collision checking is also unnecessary. 

Each expansion step of the A* algorithm is represented by a high-dimensional node, denoted as $N_{node}$, with coordinates $(x, y, z, \gamma, \beta, \alpha)$, corresponding to the object's position and attitude. The position expansion strategy for $N_{node}$ is the same as the standard A* algorithm, but for attitudes ($\gamma, \beta, \alpha$), we start the collision evaluation from the attitude closest to the parent node. If no collision occurs, we update the high-dimensional node coordinates and add them to the closed list. If a collision is detected, we choose a more deviated attitude from the parent node for re-evaluation. We pre-discretely store the shape grid for all attitudes, denoted as $M_{map}(\gamma^j, \beta^j, \alpha^j)$, capturing the object’s profile at various combinations of roll, pitch, and yaw (RPY) angles. This map is discretized to match the resolution of the overall environment map $E_{map}$.
By combining various RPY configurations, we create a multi-channel map $M_{map}$ where each channel records self-occupancy information for a specific set of discretized RPY angles.
During collision detection, a Boolean convolution operation is conducted between the environment map $M_{map}$ and the multi-channel map $M_{map}$ to efficiently assess potential collisions for each RPY configuration.
As shown in Fig. \ref{fig: front-end}, since the object's shape data is pre-loaded in memory, this collision detection process is very fast.

By prioritizing the collision-free nodes with the smallest deviations in RPY during each A* expansion, this asymmetric A* algorithm efficiently navigates high-dimensional spaces like $\mathbb{SE}(3)$.
Detailed algorithmic procedures can be found in Algorithm 2 of the supplementary materials.

The front-end produces a sequence of high-dimensional collision-free nodes that contain both position and pose information. This sequence is referred to as $\mathfrak{T}_{\textbf{A*}}$: 
\begin{align}
\mathfrak{T}_{\textbf{A*}} = \{  {N^{i}}_{node}\textbf{:}(x^i,y^i,z^i,\gamma^i, \beta^i, \alpha^i)\in \mathbb{SE}(3) \},
\end{align}
and serves as the output of the first layer in our hierarchical planning approach. 
This discrete sequence is then passed to the \textbf{mid-end}, where it is used to generate an initial trajectory. This initial trajectory, which is characterized by its speed compared to the kinodynamic approach \cite{webb2013kinodynamic}, serves as the initial input for the subsequent \textbf{back-end} optimization process.
For trajectory generation and optimization,  we employ the concept of $\mathfrak{T}{\textbf{MINCO}}$ as introduced in \cite{wang2022geometrically}. $\mathfrak{T}{\textbf{MINCO}}$ denotes the set of minimum control effort polynomial trajectories defined as follows:

\begin{align}
\begin{split}
\mathfrak{T}_{\textbf{MINCO}}=&\{p(t):[0,T_\Sigma]\rightarrow\mathbb{R}^m|\textbf{c}=\mathcal{M}(\textbf{q},\textbf{T}),\\
\textbf{q}\in&\mathbb{R}^{(N-1)m},\textbf{T}\in\mathbb{R}^M_{>0} \},  \\
\textbf{c}=&(\textbf{c}^T_1,...,\textbf{c}^T_M)^T\in\mathbb{R}^{6N\times m}, \\
\textbf{q}=&(\textbf{q}_1,...,\textbf{q}_{N-1})\in\mathbb{R}^{(N-1)\times m}, \\
\textbf{T}=&(T_1,T_2,...,T_N)^T\in\mathbb{R}^{N}\}.\\
\end{split}
\end{align}

In this representation, the trajectory $p(t)$ is an $m$-dimensional polynomial with $N$ segments, each of degree $5$. The coefficients of the polynomial are denoted as $\textbf{c}$, $\textbf{q}$ represents the intermediate waypoints. The time allocated for each segment is specified by $\textbf{T}$, with the total time denoted as $T_\Sigma = \sum_{i=1}^N T_i$. The parameter mapping $\mathcal{M}(\textbf{q},\textbf{T})$ is constructed based on Theorem 2 presented in \cite{wang2022geometrically}.

An $m$-dimensional trajectory with $M$ segments can be described as:

\begin{align}
p(t) = p_i(t - t_{i-1}) \quad \forall t \in [t_{i-1},t_i),
\end{align}

where the $i^{th}$ segment of the trajectory is represented by a polynomial of degree 5:

\begin{align}
p_i(t) = \textbf{c}^T_i \beta(t) \quad \forall t \in [0,T_i).
\end{align}

Here, $\textbf{c}_i \in \mathbb{R}^{6\times m}$ denotes the coefficient matrix, $\beta(t)=[1,t,\ldots,t^5]^T$ is the natural basis, and $T_i = t_i - t_{i-1}$ represents the time duration of the $i^{th}$ segment.

The trajectory representation in $\mathfrak{T}_{\textbf{MINCO}}$ is uniquely determined by the pair $(\textbf{q},\textbf{T})$. The mapping $\textbf{c}=\mathcal{M}(\textbf{q},\textbf{T})$ converts this representation into $(\textbf{c},\textbf{T})$, allowing for the expression of any second-order continuous cost function $J(\textbf{c},\textbf{T})$ as $H(\textbf{q},\textbf{T})=J(\mathcal{M}(\textbf{q},\textbf{T}),\textbf{T})$. Consequently, the partial derivatives $\partial H/\partial \textbf{q}$ and $\partial H/\partial \textbf{T}$ can be easily derived from $\partial J/\partial \textbf{c}$ and $\partial J/\partial \textbf{T}$.   

\textbf{Mid-End}
\label{Mid End}

Following the \textbf{front-end}, the output sequence $\mathfrak{T}_{\textbf{A*}}$ consists of discrete points without timestamps and does not constitute a trajectory. Therefore, a primary goal of hierarchical planning is to transform these reference key path points into a dynamically feasible and collision-free trajectory.

The inherent non-convexity of collision-free trajectory generation poses a significant challenge to optimization-based methods. These methods often struggle to find a global minimum and typically require well-chosen initial values, as noted in \cite{nocedal1999numerical}.
The purpose of the \textbf{mid-end} here is to provide the \textbf{back-end} with a good initial value of the optimized trajectory to reduce the optimization pressure. The trajectory generated by the \textbf{mid-end} needs to try to fit the output of \textbf{front-end}, namely the key position-pose states $(x^i,y^i,z^i,\gamma^i,\beta^i,\alpha^i) \in \mathfrak{T}_{\textbf{A*}}$ as follows:
\begin{align}
p^{i} &= (x^i,y^i,z^i),\\
R^{i} &= R_z(\alpha^i) \cdot R_y(\beta^i) \cdot R_x(\gamma^i).
\end{align}
where $R_z,R_y,R_x$ represent the rotation matrices corresponding to each axis.
To improve the alignment of trajectory path points and attitude with the desired key states from $\mathfrak{T}_{\textbf{A*}}$, we formulated an unconstrained optimization problem with the following cost function:

\begin{align}
\underset{\textbf{c},\textbf{T}}{{\min} } , \ Cost_{\textbf{mid-end}}= \lambda_m J_m+\lambda_t J_t+\lambda_p \mathcal{G}_p+\lambda_R \mathcal{G}_R.
\end{align}

Here, the terms $J_m, J_t,\mathcal{G}_p, \mathcal{G}_R$ represent the smoothness, total time, position, and pose residual penalties, respectively. The weights $\lambda_m,\lambda_t,\lambda_p, \lambda_R$ are assigned to these four items.
The position residual $\mathcal{G}p(t)$ is defined as follows, utilizing the $C^2$-smoothing function $\mathcal{L}\mu[\cdot]$:
\begin{align}
\mathcal{G}_p(t)=\mathcal{L}_\mu\left[\|p(t)-p^{i(t)}\|^2\right],
\end{align}
where $\|\cdot\|^2$ denotes the square of Euclidean norm of a vector. The function $i(t)$ maps to the index of the key node based on the proximity principle in $\mathfrak{T}_{\textbf{A*}}$, correlating the time t with the nearest key node identified in the front-end output.
The smoothing function $\mathcal{L}_\mu[x]$ is based on an exact penalty and handles non-negativity constraints, where $\mu$ is a small smoothing factor.

\begin{align}
\label{smoothedL1}
\mathcal{L}_\mu[x]= \begin{cases}0 & x \leq 0, \\ (\mu-x / 2)(x / \mu)^3 & 0<x \leq \mu, \\ x-\mu / 2 & x>\mu.\end{cases}
\end{align}

The pose residual $\mathcal{G}_R(t)$ is defined as:

\begin{align}
\mathcal{G}_R(t)=\mathcal{L}_\mu\left[\|R(t)^{-1}R^{i(t)}-I\|_F^2\right].
\end{align}

Here, $\|A\|_F^2$  represents the Frobenius norm of matrix $A$, which can be expressed as $\operatorname{tr}\left\{A^TA\right\}$ using the matrix trace.

Basically, $\mathcal{G}_R(t)$ quantifies pose similarity residuals and $\mathcal{G}_p(t)$ evaluates position similarity residuals. These optimization components are designed to ensure closeness between $p(t)$ and $p^{i(t)}$, and they strive to keep $R(t)^{-1}R^{i(t)}$ close to the identity matrix $I$. This closeness indicates the alignment of $R(t)$ with $R^{i(t)}$.

Detailed descriptions of $J_m$ and $J_t$ can be found in $\S$A of the supplementary materials. This appendix also contains gradient derivations for terms such as $\frac{\partial \mathcal{G}_{\star}}{\partial c_i  / \partial T_i}$ and $\frac{\partial J_{\star}}{\partial c_i / \partial T_i}$. It also provides insight into trajectory generation and optimization. 

After solving the mid-end unconstrained optimization problem, we construct an initial trajectory that fits the discrete pose sequence $\mathfrak{T}_{\textbf{A*}}$. This initial trajectory then serves as the starting point for further trajectory optimization. Using the SVSDF, we perform the final trajectory optimization to implement CCA while ensuring that the dynamic constraints are satisfied.

\begin{figure}[!tb] 
\centering
{\includegraphics[width=1.0\columnwidth]{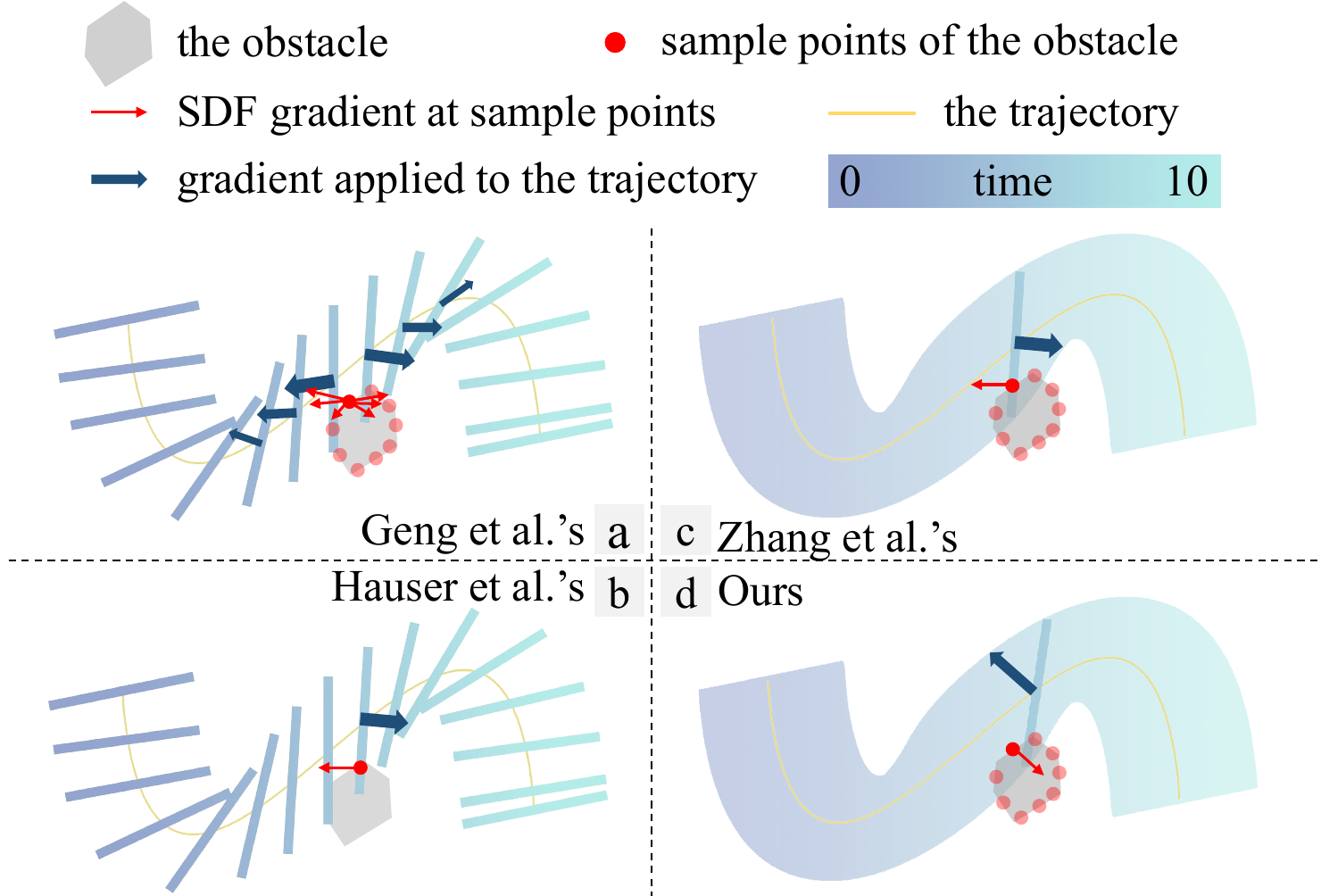}}
\captionsetup{font={small}}
\caption{ \label{fig: GSIP_grad vs traditional grad} For clarity, the figure here compares the obstacle avoidance gradients applied to the robot along the trajectory in 2D scenarios. In scenarios with more complicated robot shapes, our gradient proves to be the most effective in pushing the SV away from obstacles, thus facilitating CCA.}
\end{figure} 

\textbf{Back-End} 
\label{Back End}

The \textbf{back-end} uses the SVSDF solved by the aforementioned GSIP in Section \ref{Construction of the GSIP} to construct the optimization problem that makes the SV collision-free.
The key point is that the gradient direction provided by SVSDF is also the most appropriate direction for the continuous collision-free optimization as shown in Fig. \ref{fig: GSIP_grad vs traditional grad}. Specifically, traditional discrete sampling methods such as those described in related works \cite{wang2022geometrically,10342074,hauser2021semi}, typically compute the trajectory gradient based on the ``fixed frame state'' of the robot at some discrete time points. This approach conceptually separates the continuous motion process, making the gradient estimation at each discrete time point ``isolated''. This isolation means that the gradient estimation at each time point does not take into account the continuity and interdependence of the robot during its forward and backward motion. Such an approach ignores an important fact: the motion of robots is a continuous, holistic process, and its state at any given time is inextricably linked to its past and future states. Therefore, any approach that attempts to estimate the overall gradient by considering only information from discrete time points will not fully and accurately characterize the robot's motion in continuous space. \citet{10342104} attempted to use the gradient information of SV for trajectory optimization. However, since the SDF values inside SV are determined relative to its intrinsic shape SDF field, this method has certain limitations. Notably, the SDF values deduced inside SV are conservative, leading to erroneous gradient orientations internally. 
Our GSIP solution in Section \ref{Construction of the GSIP} provides for any obstacle point its closest distance point projection with respect to the SV. 
At the same time, we can obtain the corresponding gradient direction where the SVSDF is an accurate description of the continuous collision constraint violation in robot planning. Unlike previous works mentioned above, we derive the SVSDF accurately, especially when the obstacle is inside the SV, and no longer just get a bound inside \cite{sellan2023reach,sellan2021swept,10342104}. This allows us to get the correct gradient direction inside the SV, which facilitates continuous collision-free motion.

Using the SVSDF for collision evaluation, the cost function constructed in the \textbf{back-end} is as follows:
\begin{align}
&\underset{\textbf{c},\textbf{T}}{{\min} } \,\, \ Cost_{\textbf{back-end}}= \mathcal{G}_d+\lambda_o \mathcal{G}_o+\lambda_m J_m+\lambda_t J_t.
\end{align}
Here, the terms $J_m$ and $J_t$ represent the smoothness and total time, respectively, same to the \textbf{mid-end}.
 $\mathcal{G}_o$ and $\mathcal{G}_d$ represent the obstacle and dynamic penalty respectively, while $\lambda_o$ denotes the corresponding weight for collision avoidance. Specifically,
 \begin{align}
&\mathcal{G}_o=\sum_{i=1}^{N_{obs}}\mathcal{L}_\mu\left[J_o(\bm{x}_{ob}^i)\right],\\
\label{gradient3}
&J_o(\bm{x}_{ob})=
\begin{cases}
0, &  SVSDF(\bm{x}_{ob}) > s_{thr} ,\\
s_{thr}-SVSDF(\bm{x}_{ob}), &  SVSDF(\bm{x}_{ob})\le s_{thr}.
\end{cases}
\end{align}
where $s_{thr}$ is a safety threshold. $\bm{x}_{ob}$ is the obstacle point and $N_{obs}$ is the number of obstacle points.
Detailed explanations of gradient deviation and dynamic penalty $\mathcal{G}_d$ are provided in $\S$A of the supplementary materials.

Due to the typically non-convex shapes of objects, the exact SVSDF and its associated gradients significantly benefit the optimization process, including avoiding gradient oscillations and reducing the number of optimization iterations required. 
This has been validated in the experiments presented in Section \ref{benchmark}. Particularly in scenarios where obstacles are inside the SV, the \textbf{back-end} effectively employs the most suitable gradient provided by the SVSDF to guide the SV away from obstacles, achieving a continuous, collision-free trajectory. 
Moreover, our method can quantify the degree of collision violations between the obstacle and the SV, and rigorously determine whether the object's motion trajectory involves collisions by checking whether the computed obstacle penalty, namely $\mathcal{G}_o$, is greater than 0.

\begin{figure*}[!tb]
    \label{fig: uni_svsdf_bench}
    \centering
        \begin{minipage}{\textwidth}
        \begin{minipage}{0.24\textwidth}
         \tiny
            \centering
            \textbf{Minimum evaluation method by \citet{sellan2021swept}} 
        \end{minipage}
        \begin{minipage}{0.24\textwidth}
        \tiny
            \centering
            \textbf{GSIP model by ours} 
        \end{minipage}
        \begin{minipage}{0.24\textwidth}
        \tiny
            \centering
            \textbf{Minimum evaluation method by \citet{sellan2021swept}} 
        \end{minipage}
        \begin{minipage}{0.24\textwidth}
        \tiny
            \centering
            \textbf{GSIP model by ours} 
        \end{minipage}
    \end{minipage}
    \begin{minipage}{\textwidth}
        \begin{subfigure}{1.0\textwidth}
            {\includegraphics[width=\linewidth,height=0.3\linewidth]{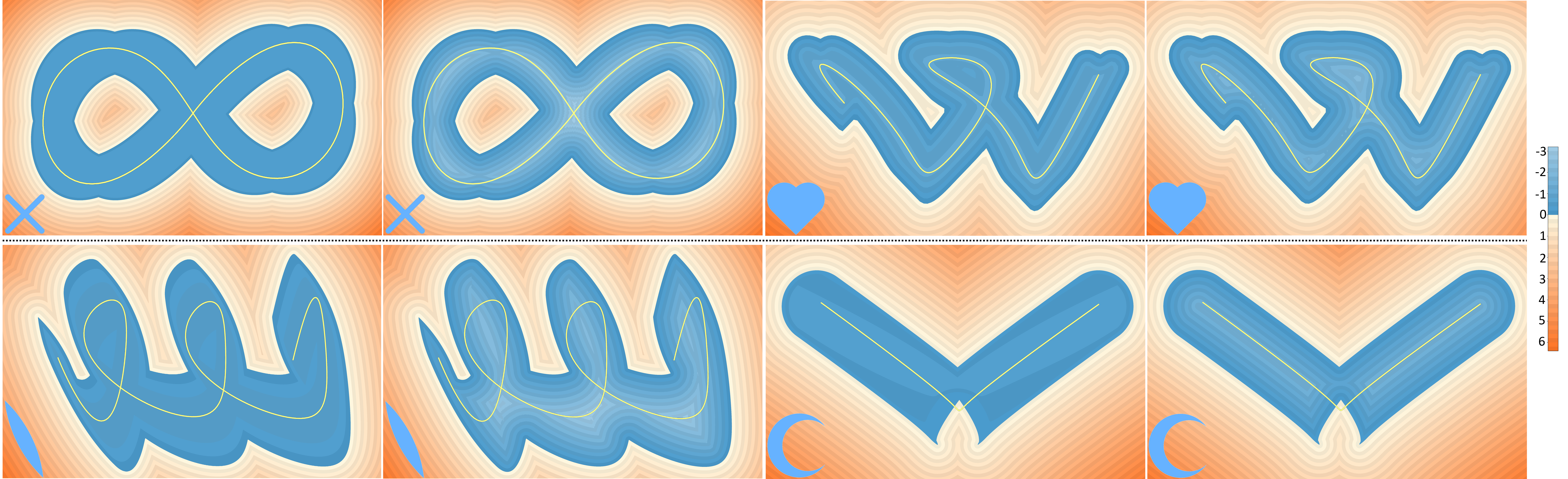}}%
        \end{subfigure}
        \caption{\label{fig: uni_svsdf_bench}This figure compares our SVSDF computation algorithm with \cite{sellan2021swept} in 2D scenarios for visual clarity. In these experiments, we move four different shapes along specific trajectories. We then show the differences in SVSDF and gradients represented by contour lines.}
    \end{minipage}
\end{figure*}
\begin{figure}[!tb] 
\centering
{\includegraphics[width=1.0\columnwidth]{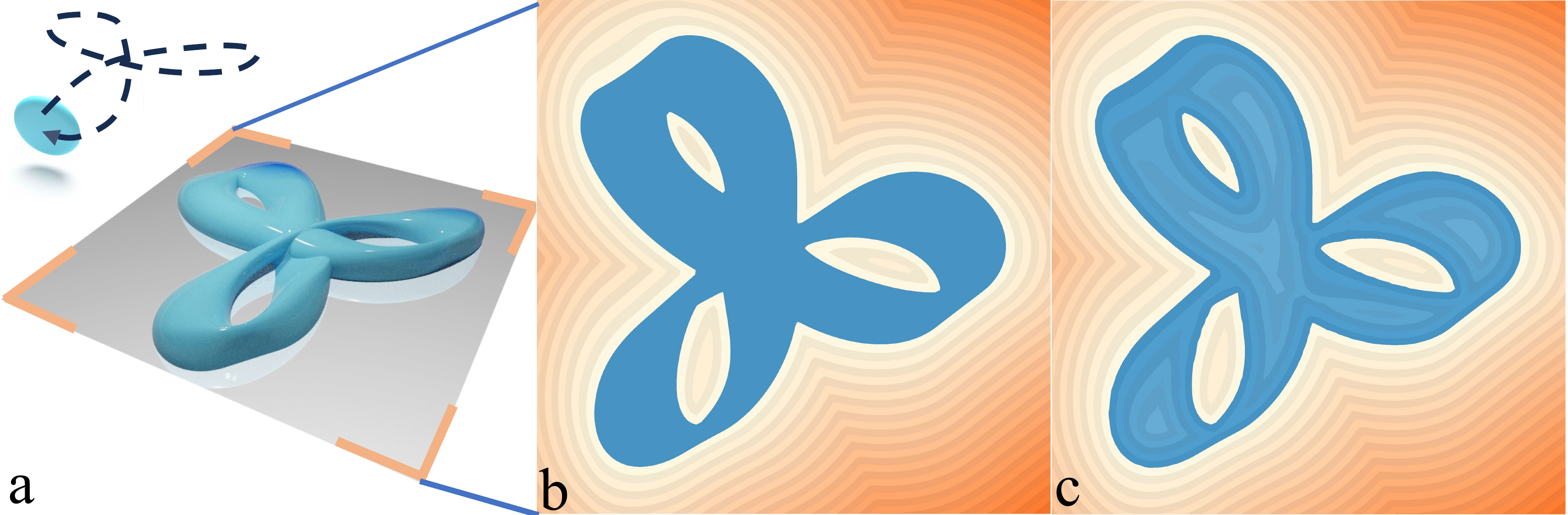}}
\captionsetup{font={small}}
\caption{ \label{3Dslice} Fig. a. shows the SV generated by a 3D round pie-shape. Unlike the internally conservative SDF  shown in Fig. b, our method, as shown in Fig. c, generates the correct SDF.}
\end{figure} 

\section{Implementation Details}
It is worth noting that the choice of tolerance, $\epsilon_{Numerical Precision}^{+}$ when solving GSIP in Alg. \ref{adaptive discretization method} does not depend on the shape of the object. In our experiments, it is set to half of the collision avoidance safety factor $s_{thr}$, which does not lead to tunneling effects. \\
\label{acceleration strategy}
\textbf{Computation of the metric function \textbf{\textit{g}}}\\
In the SVSDF calculation process, computing the metric function $g$ is a crucial step in solving the $\bm{LP(r,s)}$.
As shown in Eq. (\ref{min_distance}), the value of $g$ is the global minimum of the function $d = \mathcal{SDF}^{M(t)} \! \left( \mathcal{T}^{-1}(t) \bm{p} \right)$. Therefore, a global optimization technique is required to solve for the value of $g$. Since $d$ is a continuous function and may be non-convex, the gradient descent method can be applied to obtain a local minimum of $g$, as discussed in \cite{sellan2021swept}). It is necessary to compare several local minima to obtain the global minimum. 
In our implementation, we first substitute the original shape $M(t)$ with a bounding sphere $B$, where the function $d' = \mathcal{SDF}^{B} \! \left( \mathcal{T}^{-1}(t) \bm{p} \right)$ has an analytic expression that can be calculated with ease quickly. It is evident that $0 < d - d' < 2r$ for any $t$, where $r$ is the radius of $B$. 
Intuitively, the function $d$ lies within the band between $d'$ and $d'+ 2r$. We select intervals according to a certain time resolution and perform gradient descent in each interval to find the global minimum of $d'+2r$, $min^{B}$, and identify intervals on the function $d'$ where the value is less than $min^{B}$; thus, the global minimum of $d$ must be within these intervals. 
Through this operation, we significantly narrow down the range where the global minimum of $d$ could be located. 
Then, we further subdivide these intervals with a certain resolution and perform gradient descent again, comparing these local minima to obtain the global minimum of $d$, which is the value of $g$.

\textbf{Accelerate internal SVSDF via continuity}\\
The amplitude of SVSDF exhibits spatial continuity, which can speed up the computational process of the SVSDF inside the SV. During the iterative process of Algorithm \ref{adaptive discretization method}, the SVSDF amplitude of a neighboring point can provide an initial radius for the GSIP iteration that is close to the optimal solution. Specifically, the initial radius of the GSIP iteration can be chosen as $r_{SVSDF}^{neighbor} + d^{neighbor}$, where $d^{neighbor}$ is the distance to the neighbor point. Using this strategy to choose the initial value increases the efficiency of solving the GSIP, improving the speed by a factor of 4 to 5.


 \section{Results and evaluation}
\label{results}
Our algorithm is implemented in C++, with the trajectory optimization component relying on the L-BFGS solver \cite{liu1989limited}. 
\begin{figure}[!t] 
\centering
{\includegraphics[width=1.0\columnwidth]{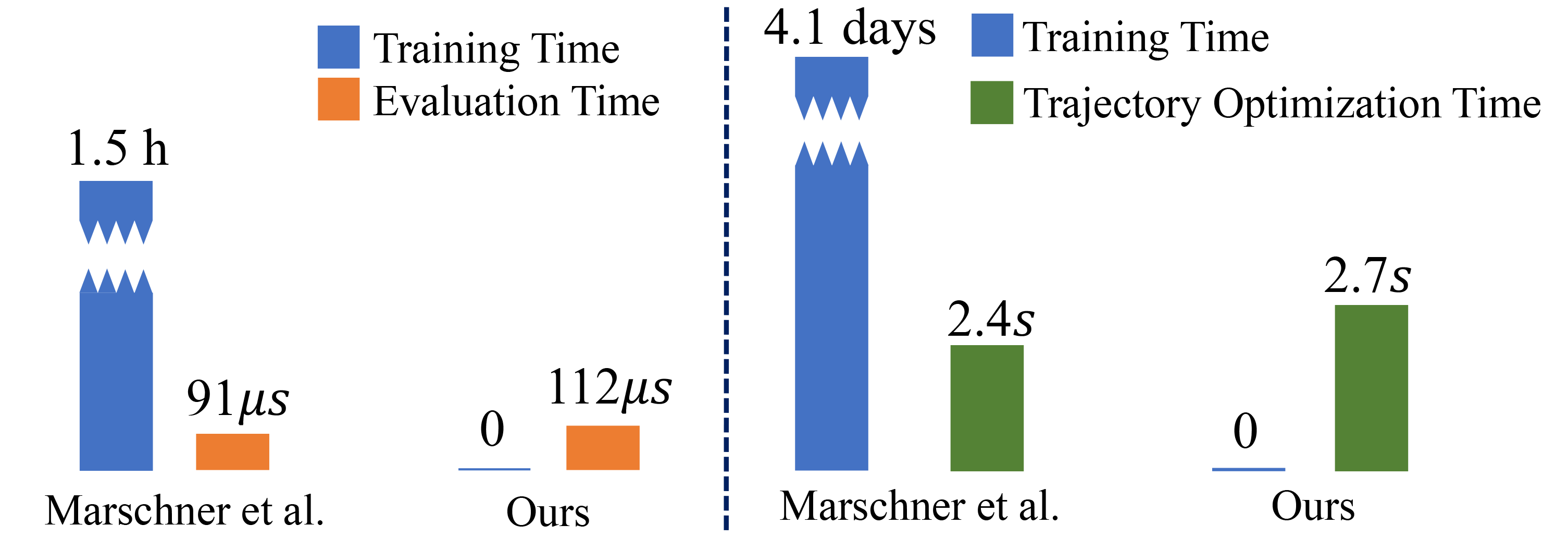}}
\captionsetup{font={small}}
\caption{ \label{fig: benchneural}  The left graph illustrates the comparative average computation time between our method and the learning-based method in calculating SVSDF for query points inside the SV. Meanwhile, the right graph shows the comparison of overall computation time for a single trajectory generation task, contrasting between our approach serving as the SVSDF computation module and the learning-based method serving the same role. The learning-based approach, which involves additional training to accommodate trajectory variations, shows significantly longer computation time. All training was performed on a single RTX2060.}
\end{figure} 
\subsection{Swept Volume SDF Results}
The first set of experiments in this section shows the results of the SVSDF computed by our method. We compared our method with the pseudo-SDF derived directly from the implicit function in \cite{sellan2021swept}.  Fig. \ref{fig: uni_svsdf_bench} and \ref{3Dslice} show cases for different shapes and trajectories. 
In contrast, our method can accurately compute signed distances and their gradients inside the SV, which plays a crucial role in trajectory optimization. The gradients computed at obstacle points inside the SV contribute significantly to guiding the SV to successfully avoid obstacles.

Additionally, we compared the average computation time of the SVSDF at one query point between our method and the learning-based method \cite{marschner2023constructive} in four example scenarios shown in Fig. \ref{fig: uni_svsdf_bench}. Although the learning-based method has parallel acceleration when processing multiple queries simultaneously, it requires pre-training to achieve sufficient accuracy. Furthermore, during the iterative process of trajectory optimization, the trajectory undergoes changes with each iteration. The learning-based method necessitates retraining after the trajectory undergoes deformation. Therefore, using this method as a component for SVSDF computation in trajectory generation tasks would result in an unacceptably long overall computation time. We conducted a trajectory generation experiment in a 2D environment with the same configuration as shown in Fig. \ref{tab:benchmarkAll}, only replacing the SVSDF computation component with the learning-based method. The statistical data are shown in Fig. \ref{fig: benchneural}.
\begin{figure*}[!tb]
   \centering
{\includegraphics[height=0.7\columnwidth]{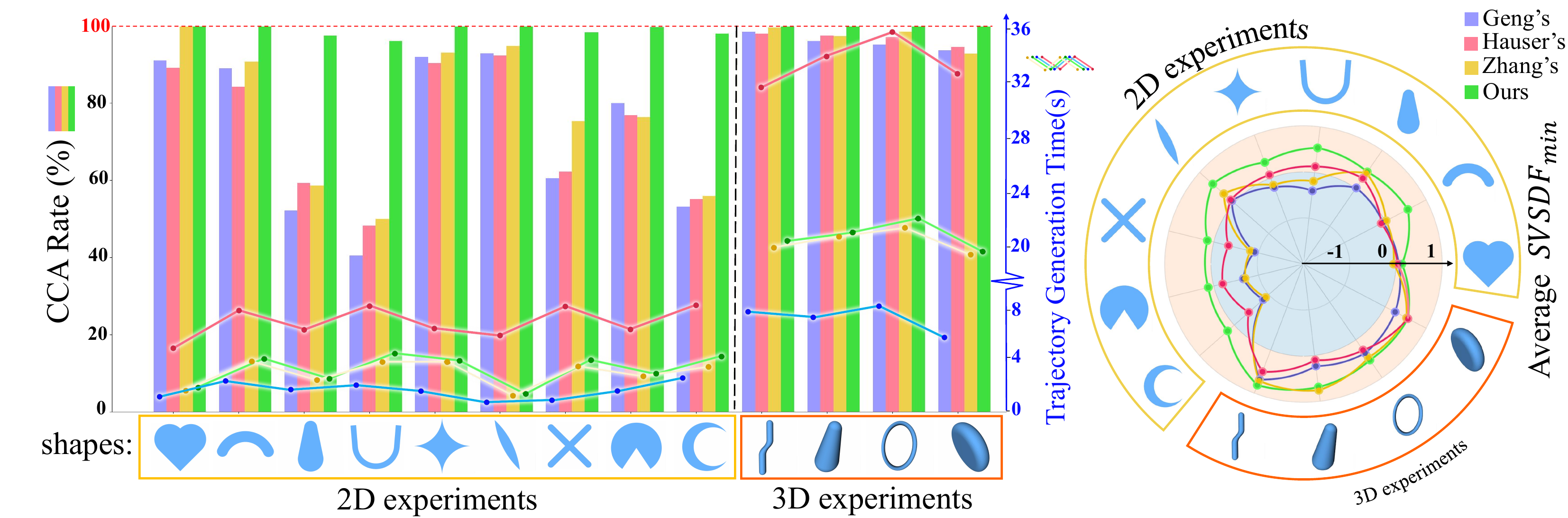}}
\captionsetup{font={small}}
\caption{ \label{tab:benchmarkAll} 
We did 500 trajectory generation tests for each shape in a randomly generated map, where the start and end points of the trajectories were randomly selected. 
The left bar graph quantifies the CCA success rates for robots of different shapes, and the line graph records the average trajectory generation time consumed. The right graph records the average minimal SVSDF values across all experiments. Given the compactness of the SV and its SDF in describing collisions, this metric serves as an effective measure for assessing the degree of CCA in trajectories.
}
\label{tab:benchmarkAll}
\end{figure*}

\subsection{Benchmarks and Ablation Study}
\label{benchmark}
We benchmark different shapes in 2D and 3D environments, evaluating their performance in two settings: environments with dense random obstacles and those with narrow gaps. These experiments demonstrate the advantages of our algorithm in CCA.
For the sake of clarity, we present two scenarios with two different shapes, as shown in Fig. \ref{fig: 2DBench}. More details on additional robot shapes can be found in Fig. 2 of the supplementary materials.
We compare our methodology with the relevant study \cite{10342074,10342104,hauser2021semi} which can handle objects of different geometries. Additionally, we conducted 500 random trials for each case and made statistical comparisons for two key metrics: the CCA success rate and the average minimal signed distance from obstacles to the SV. The CCA success rate refers to the proportion of generated trajectories that meet to be continuously collision free. The average minimal signed distance from obstacles to the SV reflects the closeness of the generated trajectories to the obstacles and serves as a more granular metric compared to the CCA success rate. The corresponding statistics are shown in Fig. \ref{tab:benchmarkAll}.

The difference in the experimental results stems from the different gradients imposed by obstacles in the optimization, as shown in the previous Fig. \ref{fig: GSIP_grad vs traditional grad}. \citet{10342074} tends to miss obstacles in complex, densely populated obstacle scenes, given its discrete collision evaluation along the trajectory and its specific gradient evaluation based on individual shapes. This characteristic makes it prone to oscillations in trajectory optimization, especially in the case of complex shapes. 
\citet{hauser2021semi} employs a branch-and-bound method to compute maximum penetration depth. However, their method also relies on evaluating gradients based on individual shapes, which does not eliminate the problem of gradient oscillation. 
The method proposed by \citet{10342104} uses continuous collision evaluation, but the presence of obstacles inside the SV during the optimization process poses significant challenges due to the incorrect signed distances and gradients imposed by these obstacles. 

The statistical results show that our method achieves the highest CCA success rate and minimizes violations of the SV collision constraints. This can be attributed to the continuity of SVSDF in collision detection and the accuracy of the internal and external signed distances and gradients.

To evaluate the individual contributions of the components of our hierarchical planner, we perform ablation experiments in 3D environments, which are detailed in $\S$B of the supplementary materials.

\begin{table}[!hb]
    \centering
    \caption{Related Parameters in Ablation Experiments and Benchmarks}
    \label{tab:parameter}
    \renewcommand\arraystretch{1.0}
    \begin{tabular}{l c c}
        \toprule
                                & Symbol  &Value      \\
\hline
   
    Max velocity ($m/s$)          & $v_m$  & 10.0 \\
    Max acceleration ($m/s^2$)          & $a_m$  & 5.0 \\
    Max jerk ($m/s^3$)          & $j_m$  & 10.0 \\
    Optimization weight for obstacles & $\lambda_o$ & 4000.0 \\
    Optimization weight for total time & $\lambda_t$   & 20.0 \\
    Optimization weight for position residuals & $\lambda_p$ & 1000.0 \\
    Optimization weight for pose residuals & $\lambda_R$ & 32000.0 \\
    Optimization weight for smoothness & $\lambda_m$ & 1.0 \\
    Optimization weight for velocity & $\lambda_v$ & 1000.0 \\
    Optimization weight for acceleration & $\lambda_a$ & 1000.0 \\
    Optimization weight for jerk & $\lambda_j$ & 1000.0 \\
    Safety threshold & $s_{thr}$ & 0.366 \\
    Discrete evaluation density in back-end &$\kappa$& 32 \\
    smoothness parameter in $\mathcal{L}\mu[\cdot]$ & $\mu$ & 0.01 \\
       \toprule
    \end{tabular}
\end{table}

Finally, we show some important parameters in Table \ref{tab:parameter} used in the above benchmark and ablation experiments.

\begin{figure*}[!tb]
     \begin{minipage}{\textwidth}
        \begin{subfigure}{0.50\textwidth}
            {\includegraphics[   width=\linewidth,height=0.6\linewidth]{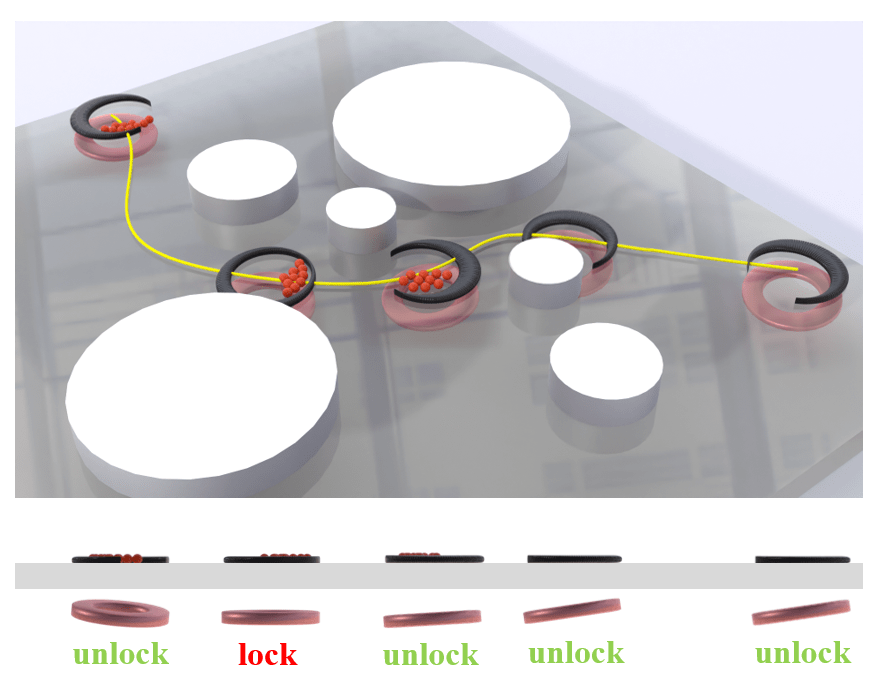}}%
          \captionsetup{font={small}}
\caption{ \label{fig: food_trans_}}
        \end{subfigure}
        \begin{subfigure}{0.50\textwidth}
             {\includegraphics[   width=\linewidth,height=0.6\linewidth]{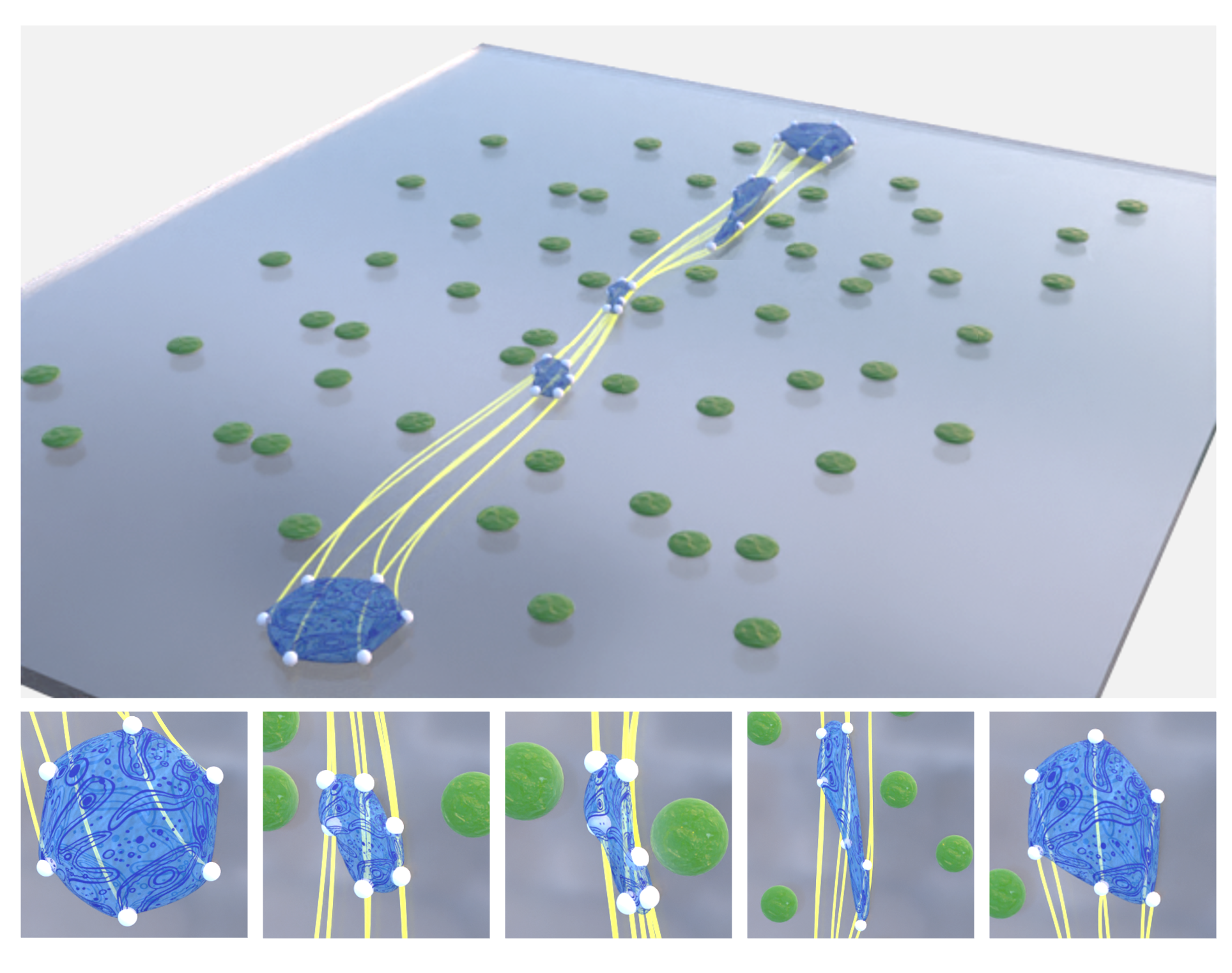}}%
           \captionsetup{font={small}}
\caption{ \label{fig: HEX_robot}}
        \end{subfigure}
\caption{Fig. a shows the planned trajectory for a deformable ferrofluid robot, driven by a magnet to transport red particles. Silver-white cylinders represent obstacles. Fig. b shows the planning result for a robot simulating deformable organisms. Each vertex of the robot is capable of omnidirectional motion.
}
\end{minipage}    
\end{figure*}

\begin{figure*}[htb]  
\centering
{\includegraphics[width=1.85\columnwidth]{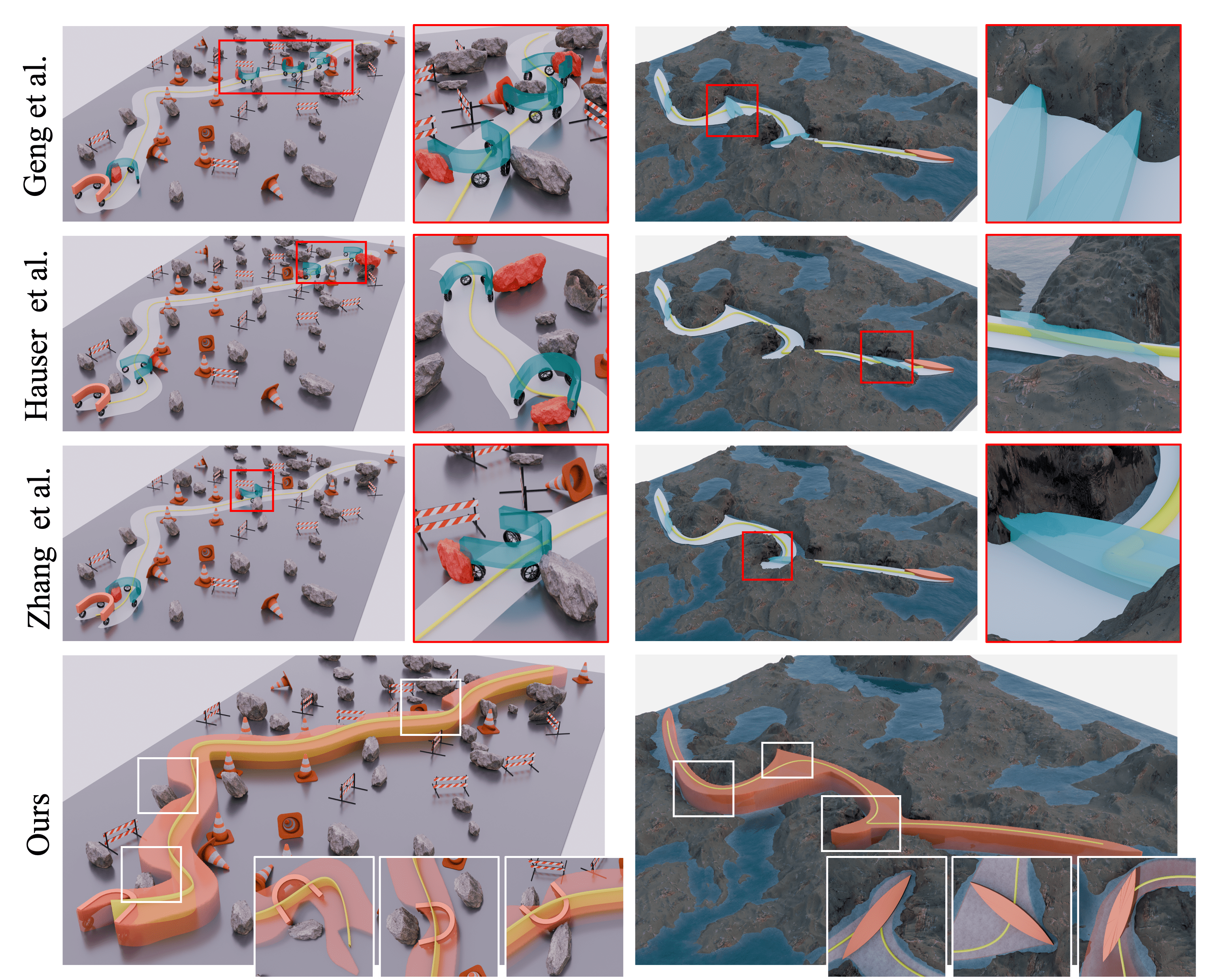}}
\captionsetup{font={small}}
\caption{ \label{fig:2DBench}
The left image shows a U-shaped car with Mecanum wheels maneuvering under a road with barriers, while the right image shows a shuttle-shaped boat navigating a narrow, rock-filled lake. 
The bottom row highlights the effectiveness of our method in CCA, showing a collision-free swept volume.
}
\label{fig: 2DBench}
\end{figure*}
\begin{figure}[!bh]  
\centering
{\includegraphics[width=0.9\columnwidth]{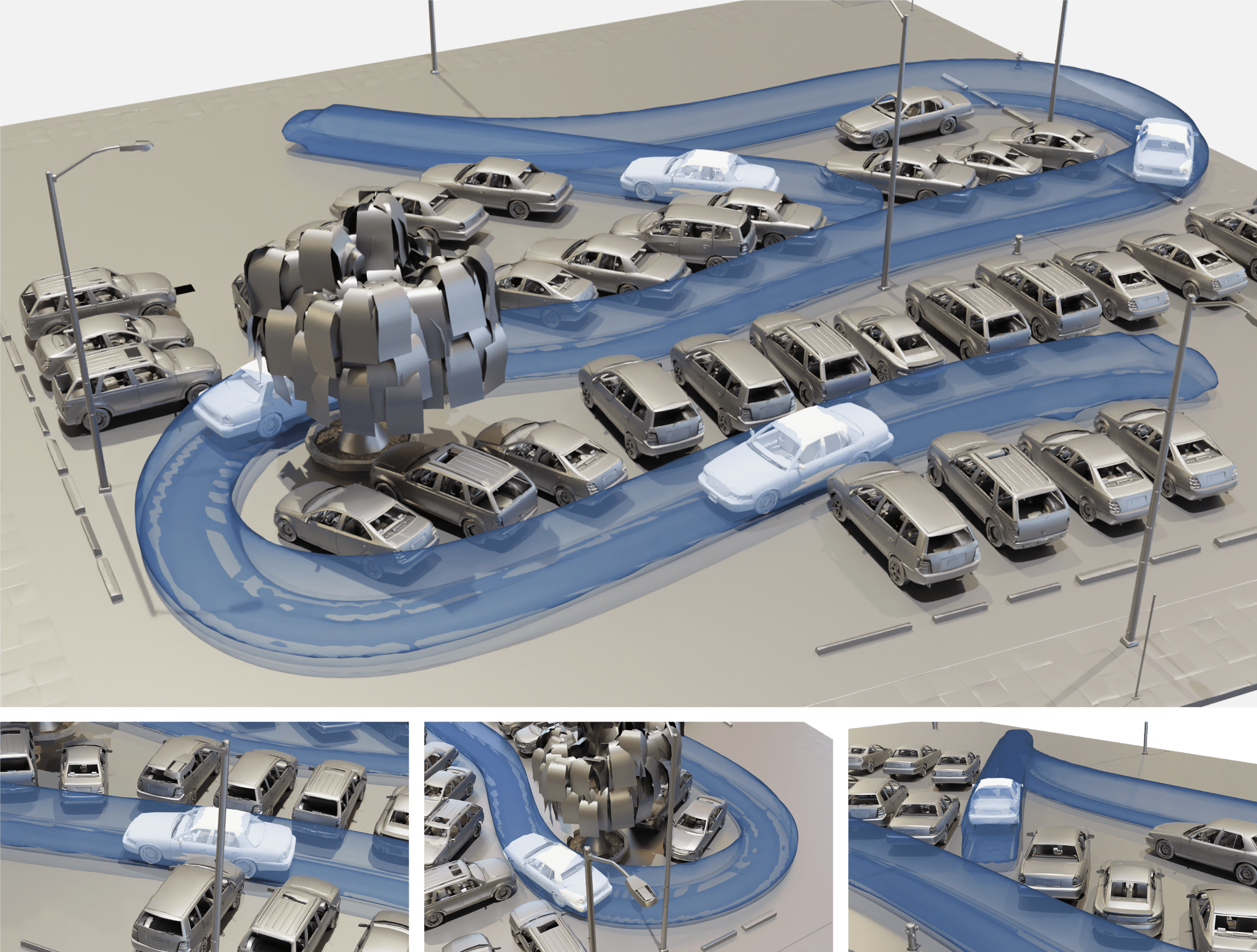}}
\captionsetup{font={small}}
\caption{ \label{fig: Car}
Utilizing the hierarchical planner, the small car is able to optimize a collision-free parking trajectory in a complex, dense environment.
}
\label{fig: Car}
\end{figure}
\subsection{Experiments}
\subsubsection{Static Shape Experiments}
\begin{figure*}[!tbh]   
\centering
{\includegraphics[width=2.0\columnwidth, height=0.88\columnwidth]{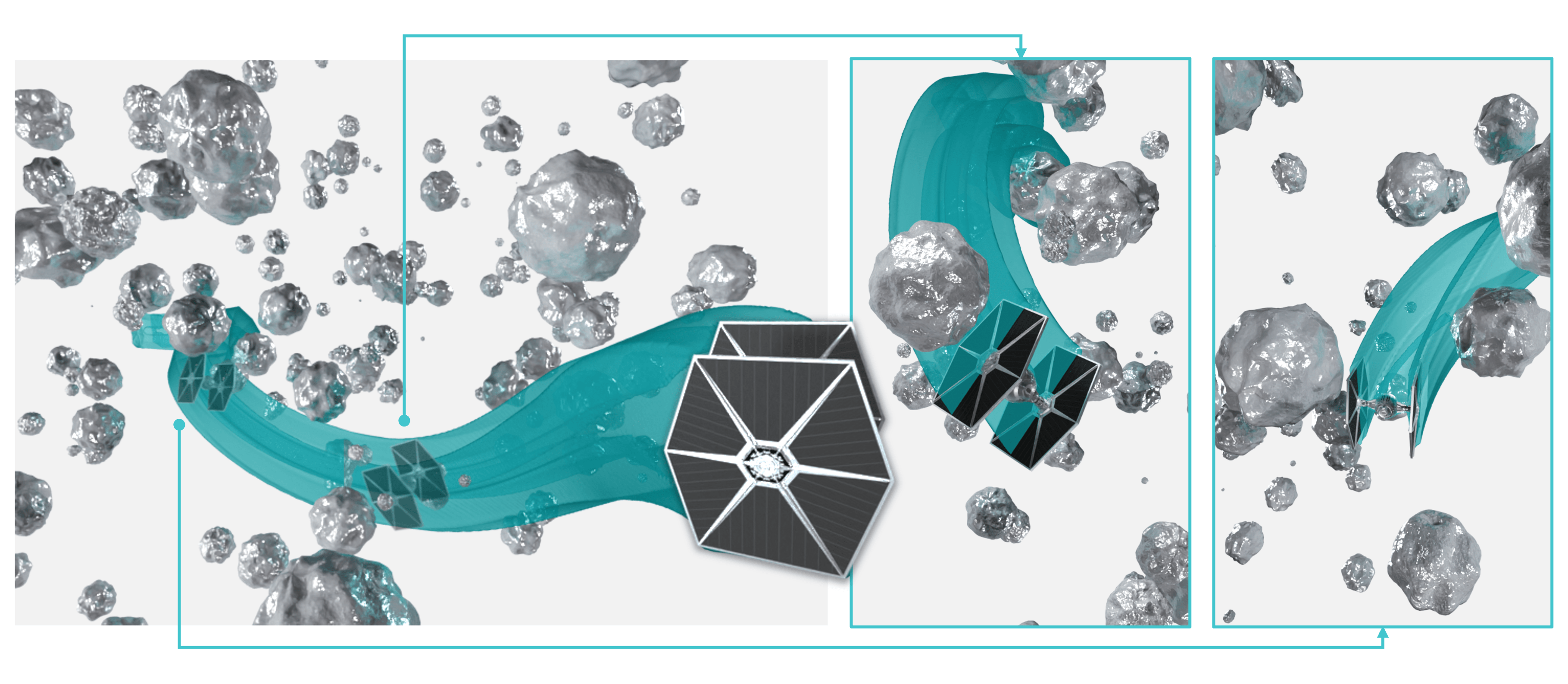}}
\captionsetup{font={small}}
\caption{ \label{fig:3DLdemo} 
The initial trajectory of the Tie fighter, along with its corresponding SV, intersects with meteorites. The complex shape of the Tie fighter challenges traditional optimization-based methods to provide optimal gradient information for obstacle avoidance. However, our hierarchical planning framework, especially the SVSDF-based backend, ensures that the final optimized SV is collision-free.} 
\label{fig: 3DLdemo}
\end{figure*}

\begin{figure}[!h]   
\centering
{\includegraphics[width=1.0\columnwidth]{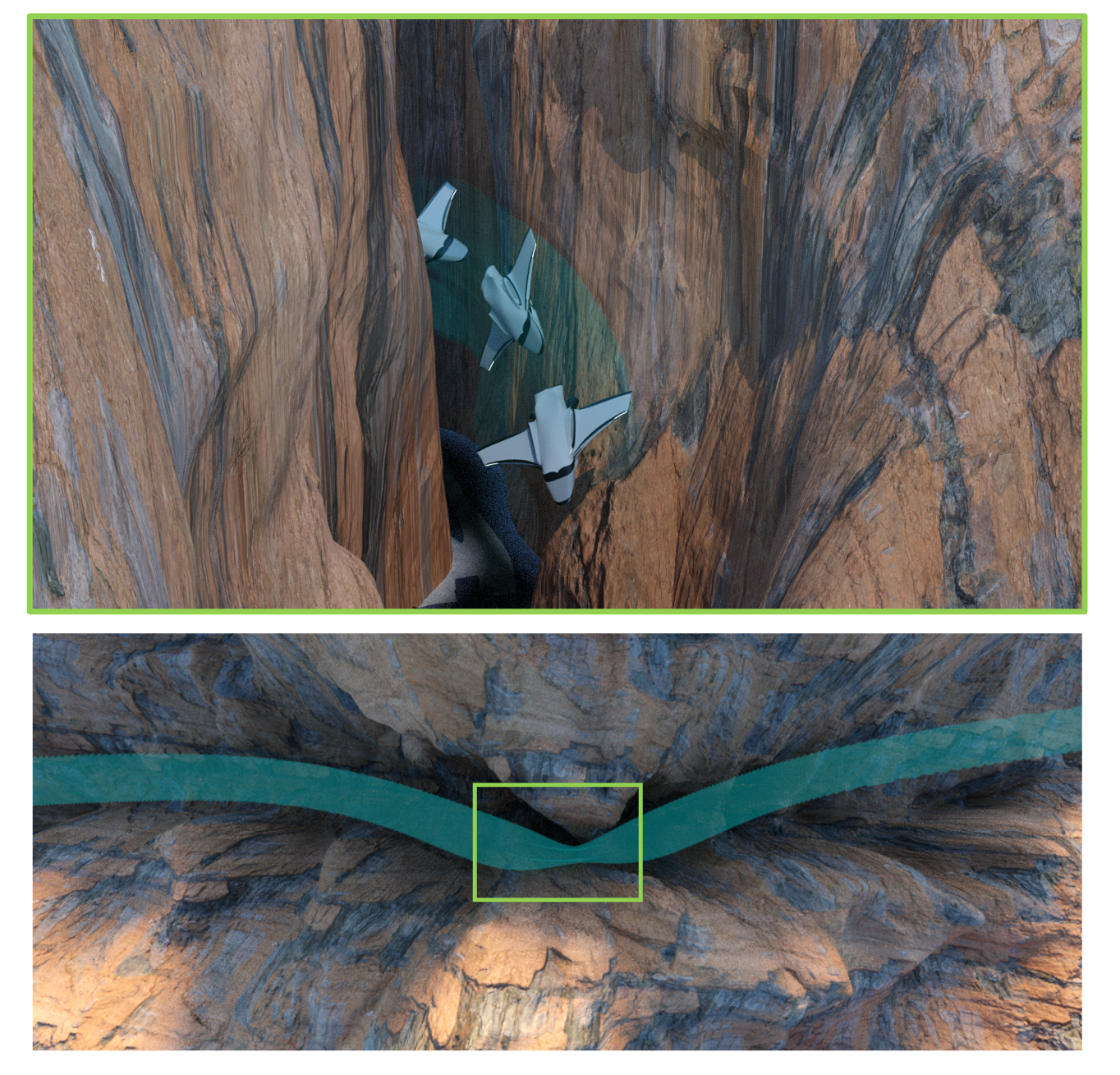}}
\captionsetup{font={small}}
\caption{ \label{fig:space_ship}
Owing to the precise collision representation in the SVSDF, the spacecraft efficiently plans continuous, collision-free trajectories in extremely confined canyon spaces. This performance exceeds that of planning methods that rely on discrete collision evaluation.}

\label{fig: gorge}
\end{figure}

In Fig. \ref{fig: 3DLdemo}, we simulate a TIE fighter navigating through a complex environment in space filled with asteroids. Our method generates a smooth, continuous collision-free trajectory that is consistent with the dynamic model. Notably, we use the original mesh model of the TIE fighter without any specific simplifications. In this scenario, we use the multirotor dynamic model \cite{faessler2017differential}. 
In fact, by altering the formulas in the optimization process, our approach can be adapted to various dynamic models. Fig. \ref{fig:space_ship} demonstrates our method in planning a flight trajectory for a fixed-wing aircraft navigating through an extremely narrow canyon, following the dynamic model of the fixed-wing aircraft. Our approach is also applicable to autonomous driving. Fig. \ref{fig: Car} illustrates how our method facilitates a car to perform seamless, collision-free automated parking in a densely packed parking lot. These cases effectively extend the work on ``Path planning'' discussed in \cite{sellan2021swept}. While \citet{sellan2021swept} only reconstructed the surface of the SV, we used our pipeline to achieve CCA in planning for robots of various shapes.
Fig. \ref{fig:siggraphwalldemo} highlights the feature that our method does not sacrifice any solution space.
In this experiment, objects with the shapes of the characters ``SIGGRAPH'' and the logo pass through three walls with holes. Our method successfully generates continuous collision-free trajectories, even though the shapes of the holes are almost identical to those of the flying objects.

It should be emphasized that the SVs depicted are used for visualization purposes only. Actually, our method does not require the explicit reconstruction of SV surfaces. The technique for generating SV is based on this work: \cite{sellan2021swept}.
\subsubsection{Deformable Shape Experiments}
Our method is also effective for deformable shapes, requiring only that the shape's change, $M(t)$, be differentiable. This section presents novel scenarios to demonstrate the effectiveness of our method in generating trajectories for such shapes.
As a first example, we consider the crescent-shaped ferrofluid robot that transports particles (inspired by \cite{fan2020reconfigurable}).
The ferrofluid robot can deform between crescent and annular shapes by adjusting the tilt angle of the annular magnet beneath it.
As shown in Fig. \ref{fig: food_trans_}, the robot successfully avoids obstacles and transports the red particle.
In our second example, we conceptualize a robot model inspired by ``shape-shifting organisms''. This model consists of vertices that can move independently, allowing the polygon formed by these vertices to have a high degree of deformability. This design holds promise for a variety of applications in various domains, including flexible multi-robot collaborative transportation. In our approach, we treat the trajectories of individual vertices as optimization objectives. We use the polygon defined by these vertices as the shape of the robot and apply our trajectory generation algorithm. The results of this process are shown in Fig. \ref{fig: HEX_robot}.
\section{Conclusion and Limitations}
To the best of our knowledge, by solving the GSIP model, our algorithm is the first non-deep-learning algorithm to calculate the exact SVSDF of arbitrary shapes. Our pipeline integrates computational techniques of SV from computer graphics with trajectory optimization techniques from robotics, combining these two fields to provide an innovative and effective solution for CCA navigation for robots of various shapes in complex environments. 
In addition, our method can also be applied to interpolation animation generation, volume representation/rendering, reverse engineering, physics simulation, and CAD/CAM fields. 
We are committed to open-source our algorithm for the benefit of the community.

Our algorithm also has its limitations. First, due to the strong non-convexity of the trajectory optimization problem, although constructing safety constraints using an exact SVSDF can effectively improve CCA metrics, it is still not guaranteed to achieve $100\%$ CCA. 
Second, in a 3D environment, computing SVSDF requires extensive evaluation computations, resulting in non-real-time trajectory generation. Therefore, we are actively exploring the use of spatio-temporal continuum techniques to further optimize the computational speed.  Third, our current approach uses sample points to represent obstacles. A possible direction for improvement is to extend the concept of SDF by computing the distance from an object to the nearest obstacle in $\mathbb{SE}(3)$ space.
Fourth, our method is not well suited to handle dynamic obstacles. Our approach to dynamic obstacles involves treating the SV of the obstacle as a static obstacle. 
A possible extension is to obtain the relative trajectory by computing the difference between the trajectories of the object and the obstacle. This leads to an extension of the concept of ``swept volume'' to ``relative motion swept volume'', a direction we intend to explore in the future.
\begin{figure}[!h]  
\centering
{\includegraphics[width=1.0\columnwidth]{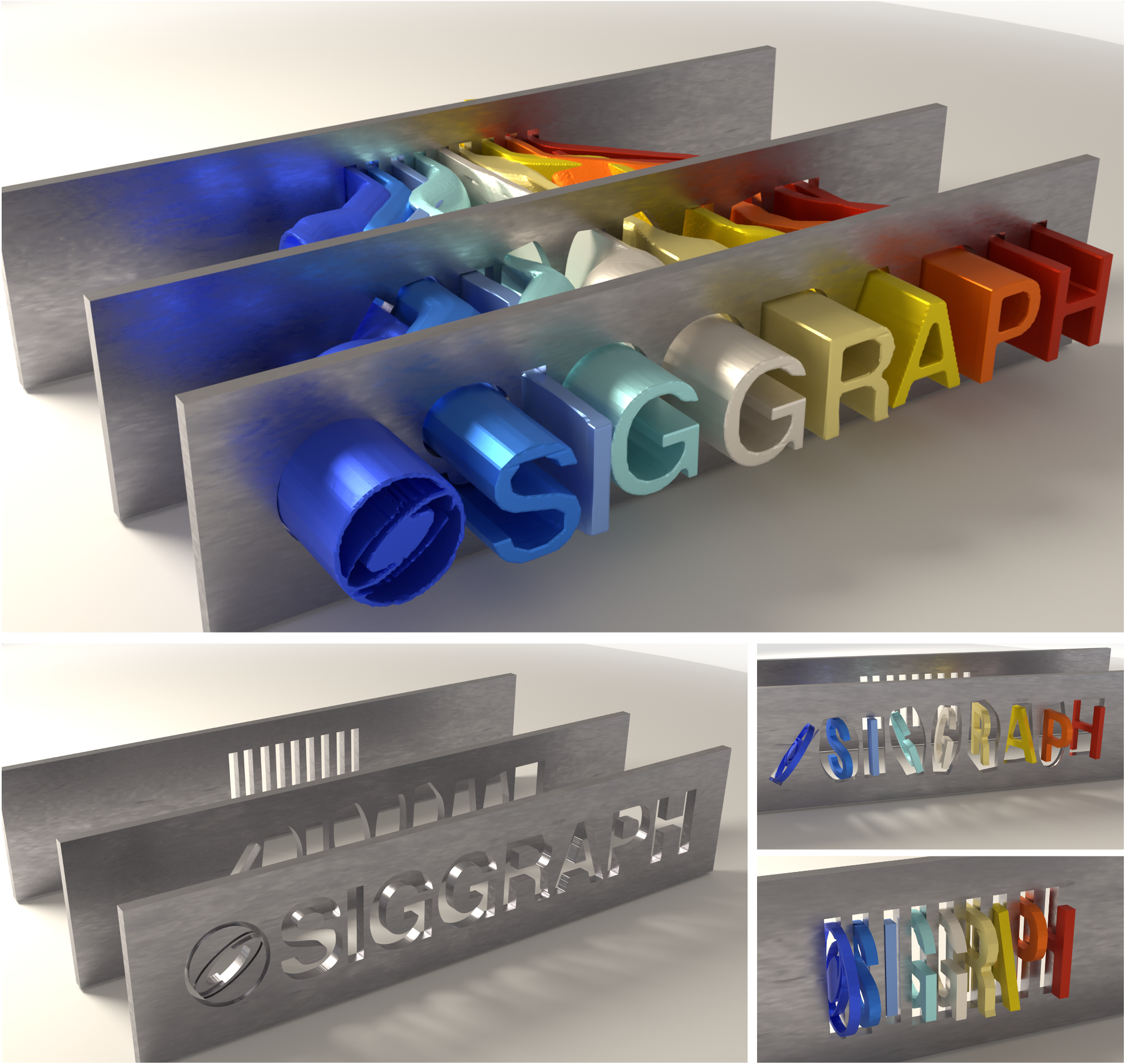}}
\captionsetup{font={small}}
\caption{ \label{fig:siggraphwalldemo}The ``SIGGRAPH'' logo and letters navigate continuously and collision-free through the narrow gaps formed by three walls.}
\label{fig:siggraphwalldemo}
\end{figure}
\clearpage
\clearpage 
\bibliographystyle{ACM-Reference-Format} 
\bibliography{sample-base}
\end{document}